\pdfoutput=1
\documentclass{article}

\PassOptionsToPackage{numbers, compress}{natbib}

\usepackage[final]{neurips_2024}

\usepackage[utf8]{inputenc} 
\usepackage[T1]{fontenc}    
\usepackage[linkbordercolor={0.9 0.8 1}]{hyperref}

\usepackage{url}            
\usepackage{booktabs}       
\usepackage{multirow}
\usepackage{amsfonts}       
\usepackage{nicefrac}       
\usepackage{microtype}      
\usepackage[table]{xcolor}
\usepackage[pdftex,final]{graphicx}
\usepackage{wrapfig}
\usepackage{subcaption}
\usepackage{array}
\usepackage{adjustbox}
\usepackage{caption}
\usepackage{amsmath}
\usepackage{algorithm}
\usepackage{algpseudocode}
\newcommand{\as}{\vspace{0.3em}}
\renewcommand{\algorithmicindent}{0.5em}
\usepackage[colorinlistoftodos]{todonotes}
\newcommand{\SCAP}{{SCAP}}
\newcommand{\texttau}{$\tau$ }

\title{Post-Training Statistical Calibration \\for Higher Activation Sparsity}

\author{%
  Vui Seng Chua\thanks{equal contribution}
  \And Yujie Pan\footnotemark[1]
  \And Nilesh Jain
  \AND
  \\
    Intel Corporation\\
    \texttt{\{vui.seng.chua; yujie.pan; nilesh.jain\}@intel.com}
}

\begin{document}
\maketitle

\begin{abstract}
We present Statistical Calibrated Activation Pruning (SCAP), a post-training activation pruning framework that (1) generalizes sparsification by input activations of Fully-Connected layers for generic and flexible application across Transformers, and (2) features a simple Mode-Centering technique to pre-calibrate activation distributions for maximizing post-training sparsity. Our results demonstrate robust Pareto efficiency compared to prior methods, translating to a 1.5$\times$ additional LLM decoding speedup against CATS\cite{lee_cats_2024} at iso model quality. SCAP effectiveness is empirically verified across a wide range of models, including recent Transformer Decoders, MoE, Mamba2, Encoding Transformer, and pre-quantized models, highlighting its practicality and scalability. The code is available \href{https://github.com/IntelLabs/SCAP}{here}.
\end{abstract}

\section{Introduction}
Activation sparsity is an emerging model optimization method for efficient deployment of Large Language Models (LLMs). Extensive studies in \emph{Lazy Neuron Phenomenon}\cite{li_lazy_2023} reveal a high prevalence of activation sparsity at the output of ReLU after pretraining, in both encoder or decoder Transformers, across different model sizes and tasks, including language and vision. Intriguingly, larger models tend to exhibit greater sparsity. Deja Vu\cite{liu_deja_2023} coined the term of \emph{contextual sparsity} and uncovered that, along with the feed-forward networks (FFN), sparsity also exists within the activation of attention layers on a per-input basis. To utilize this sparsity for inference efficiency, sparse neuron predictors \cite{liu_deja_2023, akhauri_shadowllm_2024} have been introduced to dynamically forecast and skip the redundant operations in attention heads and FFNs. \cite{song_powerinfer_2023, alizadeh_llm_2024} further leverage the locality of sparse neurons to develop efficient weight partitioning strategies, addressing the memory challenges when deploying LLMs on consumer CPU-GPU systems. 

\begin{figure}
\begin{minipage}{0.5\textwidth}
\centering
\small
\resizebox{0.75\textwidth}{!}{%
\begin{tabular}{llc}
\toprule
{ LLM family} & { \begin{tabular}[c]{@{}l@{}}Release \\ Year/Month\end{tabular}} & { \begin{tabular}[c]{@{}l@{}}Activation\\ Function\end{tabular}} \\ \midrule
{ T5}         & { 2019/10}                                                       & { \textcolor{cyan}{ReLU}}                                                          \\
{ GPT3}       & { 2020/06}                                                       & { GELU}                                                          \\
{ PaLM}       & { 2022/04}                                                       & { SiLU}                                                          \\
{ OPT}        & { 2022/05}                                                       & { \textcolor{cyan}{ReLU}}                                                          \\
{ BLOOM}      & { 2022/07}                                                       & { GELU}                                                          \\
{ Llama}     & { 2023/02}                                                       & { SiLU}                                                          \\
{ Falcon}     & { 2023/05}                                                       & { GELU}                                                          \\
{ MPT}        & { 2023/05}                                                       & { GELU}                                                          \\
{ Llama2}     & { 2023/07}                                                       & { SiLU}                                                          \\
{ Mistral}    & { 2023/09}                                                       & { SiLU}                                                          \\
{ ChatGLM3}   & { 2023/10}                                                       & { SiLU}                                                          \\
{ Gemma}      & { 2024/02}                                                       & { GELU}                                                          \\
{ Llama3}      & { 2024/04}                                                       & { SiLU} \\
{ Qwen2}      & { 2024/06}                                                       & { SiLU} \\
{ Gemma2}      & { 2024/06}                                                       & { GELU} \\
{ Mamba2}      & { 2024/06}                                                       & { SiLU} \\
{ Phi3.5}      & { 2024/08}                                                       & { SiLU} \\
\bottomrule
\end{tabular}
}
\captionsetup{hypcap=false}
\captionof{table}{Decline usage of ReLU in recent LLMs}

\label{tab:llm_act}
\end{minipage}
\hfill
\begin{minipage}{0.475\textwidth}
    \centering
    \includegraphics[width=1\linewidth]{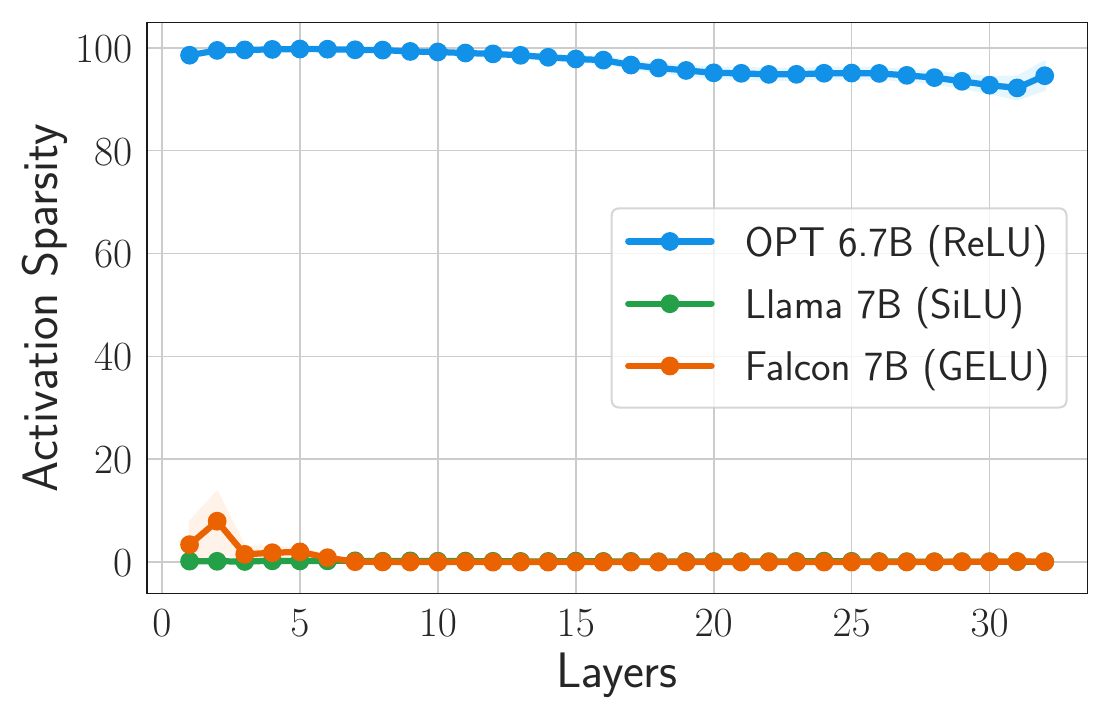}
    \caption{ReLU output is sparse while SiLU and GELU outputs are dense. Extracted from \cite{mirzadeh_relu_2023}.}
    \label{fig:actfunc_sparsity}
\end{minipage}
\end{figure}

While sparse activation can be exploited for accelerating inference, prior methods hinge on the inherent sparsity of ReLU, which presents a challenge as ReLU has fallen out of favor in recent LLM families (see Table \ref{tab:llm_act}). Due to their greater training convergence\cite{shazeer_glu_2020}, SiLU and GELU have seen increased adoption, prompting new methods to induce sparsity in their dense activations (see \autoref{fig:actfunc_sparsity}). \emph{Relufication}\cite{mirzadeh_relu_2023} advocates the reinstatement of ReLU as the primary activation function and as a means for pruning activations within the LLMs. Variants of Relufication have been explored to maximize sparsity on non-ReLU LLMs. ReLU$^{2}$\cite{zhang_relu2_2024} chains two ReLUs as the activation function while dReLU\cite{song_turbo_2024} discards the SiLU but places ReLU at the output of Up and Gate projections, respectively in the GLU-based FFN. Despite attaining high sparsity, the representation of Relufied LLMs is significantly disrupted, necessitating extensive full-model uptraining and advanced recipes \cite{song_prosparse_2024} to restore the  model’s capabilities. This process demands cluster-level compute resources, which are often inaccessible to many. Along with prolonged turnaround times, Relufication inflates costs and limits the scalability of LLM deployment.

In contrast, post-training optimization offers a simpler and more cost-effective approach, as exemplified by weight quantization methods such as GPTQ\cite{frantar_gptq_2023} and AWQ\cite{lin_awq_2024}. These techniques determine quantization parameters through calibration with a small set of text prompts on pretrained or instruction-tuned LLMs, making them highly efficient and potentially allowing the optimization to be executed directly on the target deployment device rather than computing clusters. 

To our knowledge, CATS\cite{lee_cats_2024} represents the first instance of post-training activation sparsification in the LLM era. Similar to the magnitude-based weight pruning \cite{han_learning_2015, sun_simple_2024}, CATS masks out post-SiLU activations of  Mistral-7B and Llama-7B based on a calibrated threshold. Unlike the rigidly sparse Relufied LLMs, CATS outperforms Relufication by demonstrating that activation sparsity of the GLU-FFNs can be controlled to trade off downstream task performance within 1-2\% of the original models, all without any additional training.

In this early work, we extend the generality of the post-training activation pruning framework with a statistical pre-processing technique, which we refer to as \textbf{Statistical Calibrated Activation Pruning (\SCAP)}. Our contributions include: 
\begin{enumerate}
    \item \textbf{Generalized Activation Pruning for Pareto Efficiency:} SCAP proposes to sparsify input activations of Fully-Connected (FC) layers, entailing a universal pruning and kernel implementation across all FCs within Transformer architectures. This approach goes beyond conventional post-activation sparsification, allowing for flexible sparsity across different layers without requiring additional predictor training or custom inference patterns. We demonstrate that SCAP surpasses the prior post-training CATS method in both accuracy and sparsity, achieving a more optimal trade-off between computational efficiency and task performance. Notably, with only a -1.5\% deviation from the baseline on Mistral-7B across a set of zero-shot tasks, SCAP attains 48.5\% FFN sparsity compared to CATS’ 33.3\%, translating to a 1.5$\times$ additional speedup in decoding latency over CATS, a 27.1\% overall improvement against the dense model.
    \item \textbf{Mode-Centering for Enhanced Sparsity}: We empirically observed that skewed and shifted activation distributions, artifacted by preceding layer, limit prunability. To address this, we introduce a novel Mode-Centering pre-calibration technique that estimates the mode of an activation distribution and shifts it to zero while preserving computation outcomes. This approach effectively maximizes sparsity opportunities for $L_1$-based pruning, particularly in the Down projection of non-GLU FFNs. Empirical results show up to a 44.7\% increase in input sparsity with negligible loss in performance of MPT-7B, along with a substantial increase in Falcon-7B and Vision Transformer.
    \item \textbf{Extensive Model Coverage:} We showcased the applicability of SCAP to a wide range of models, including Transformer Decoders, MoE, Mamba2 and Vision Encoding Transformers, as well as pre-quantized models (see \autoref{tab:many_models}). This highlights the advantages of SCAP in terms of turnaround time and scalability, emphasizing the practicality of post-training activation sparsification.
\end{enumerate}

\section{Generalized Post-Training Activation Pruning}
\label{sec:generalized_ptap}
The premise of acceleration through sparsity lies in the elimination of ineffectual operations involving zeros. During the decoding phase of LLMs, each newly generated token serves as input for subsequent forward pass, necessitating matrix-vector multiplications (GEMV) at every fully-connected (FC) layer. A sparse input vector (activation) forms a dot product with a correspondingly structured sparse matrix, leading to computational efficiency. The resultant dot product reduces memory bandwidth requirements by eliminating the need to fetch sparse weight channels, a critical improvement since memory bandwidth is the primary bottleneck during the decoding phase\cite{pope_efficiently_2023}. It also reduces multiply-add-accumulate compute cycles along the sparse inner dimensions.

\textbf{Challenges of reliance on post-activation sparsity} often involves intricate execution schemes to maximise sparse execution. For instance, CATS\cite{lee_cats_2024} adopts post-SiLU activation pruning. To attain higher sparsity in the GLU-based FFNs, it is necessary to compute the Gate projection, followed by SiLU and pruning operator in advance to identify redundant output channels in the Up projection. However, this approach may be suboptimal when the Up and Gate projections are consolidated into a single execution, or in cases such as parallel Attention-FFN architectures\cite{chowdhery_palm_2022, wang_gptj, almazrouei_falcon_2023}, where Query, Key, Value, Up, and Gate projections are typically fused for efficiency\cite{pope_efficiently_2023}. An alternate approach is to employ predictors \cite{liu_deja_2023, song_powerinfer_2023, akhauri_shadowllm_2024,song_turbo_2024} to estimate \textit{a priori} the locations of sparse neurons in the post-activation, thereby avoiding unnecessary output channels in the Up projection.

Contrary to recent activation pruning techniques, which predominantly target the output of activation functions in FFNs, we propose \textbf{a generalization by sparsifying only the input activations to FC layers}. This approach enables a unified calibration process and a generic sparse kernel implementation across any FC layers within Transformers, including those in attention blocks. It decouples targeted FCs, allowing for a flexible combination of input sparsities, resulting in more and sparser FCs for greater acceleration. In addition, direct pruning on the input activation of the Up/Gate projection also eliminates the additional cost of training predictors, streamlining the optimization process, as well as reducing the inference overhead associated with runtime predictions.

\begin{figure}[htbp]
    \centering
    \begin{subfigure}[b]{0.24\textwidth}
        \centering
        \includegraphics[width=\textwidth]{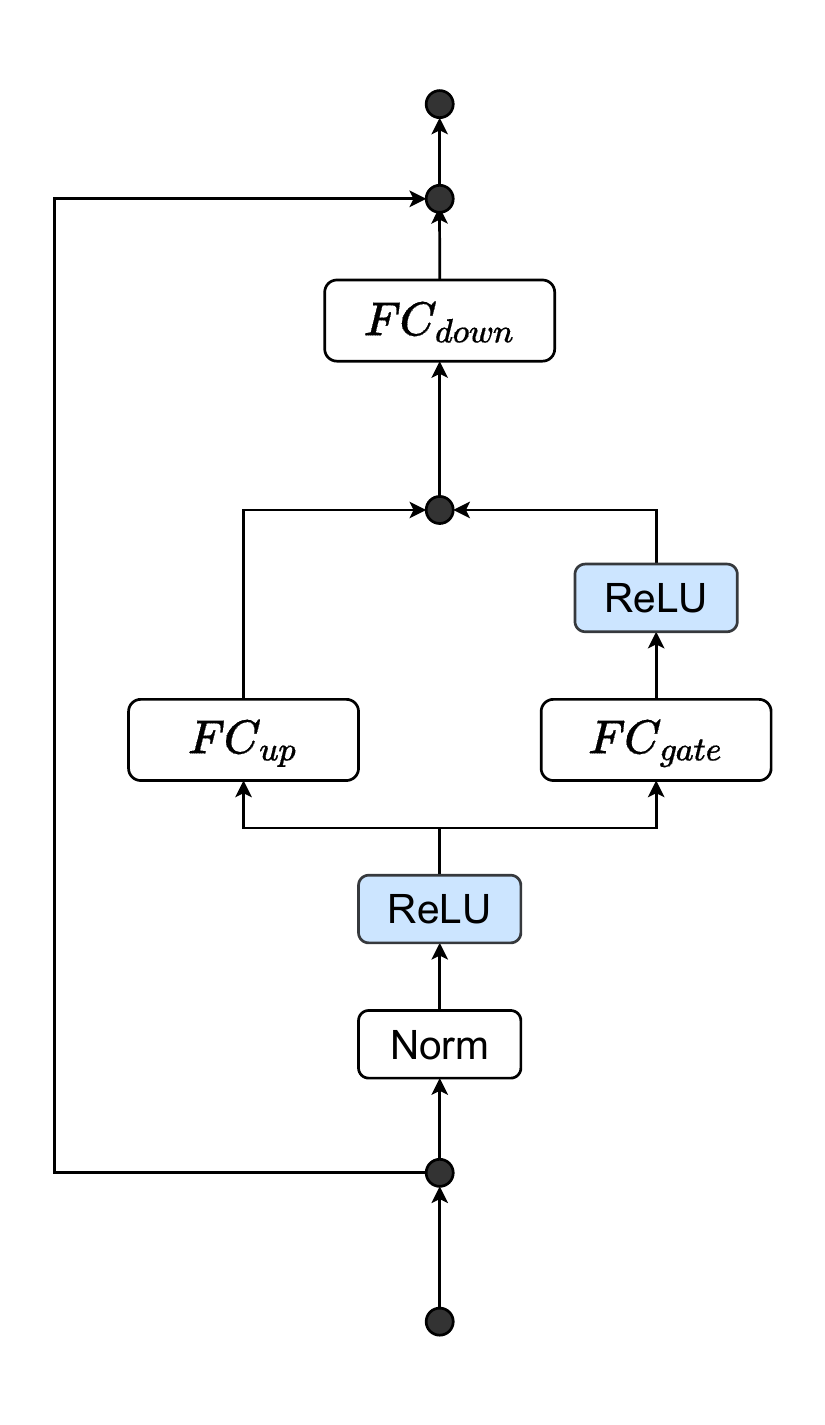}
        \caption{Relufication \cite{mirzadeh_relu_2023}}
        \label{fig:swiglu_relufication}
    \end{subfigure}
    \hfill
    \begin{subfigure}[b]{0.24\textwidth}
        \centering
        \includegraphics[width=\textwidth]{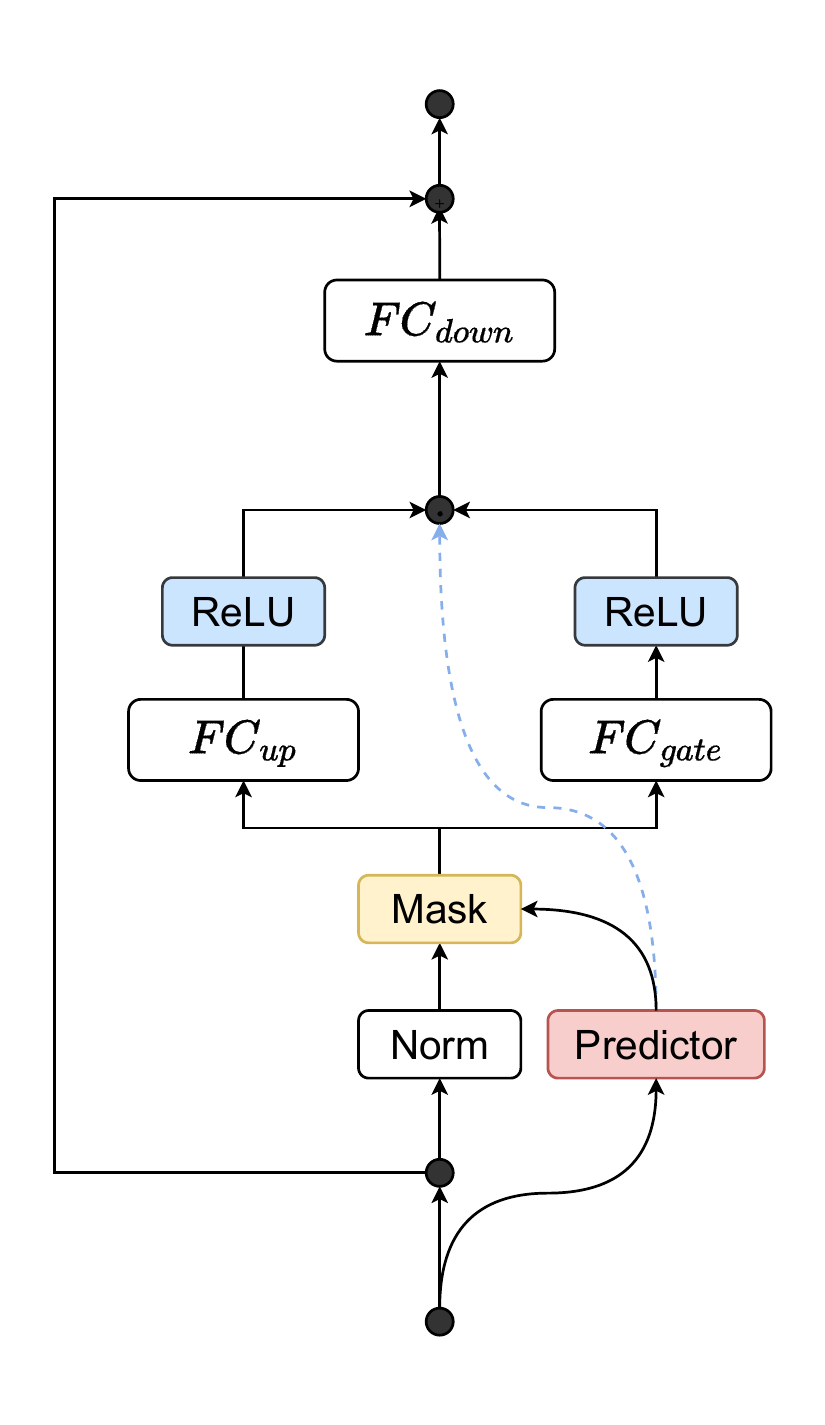}
        \caption{TurboSparse \cite{song_turbo_2024}}
        \label{fig:swiglu_turbosparse}
    \end{subfigure}
    \hfill
    \begin{subfigure}[b]{0.24\textwidth}
        \centering
        \includegraphics[width=\textwidth]{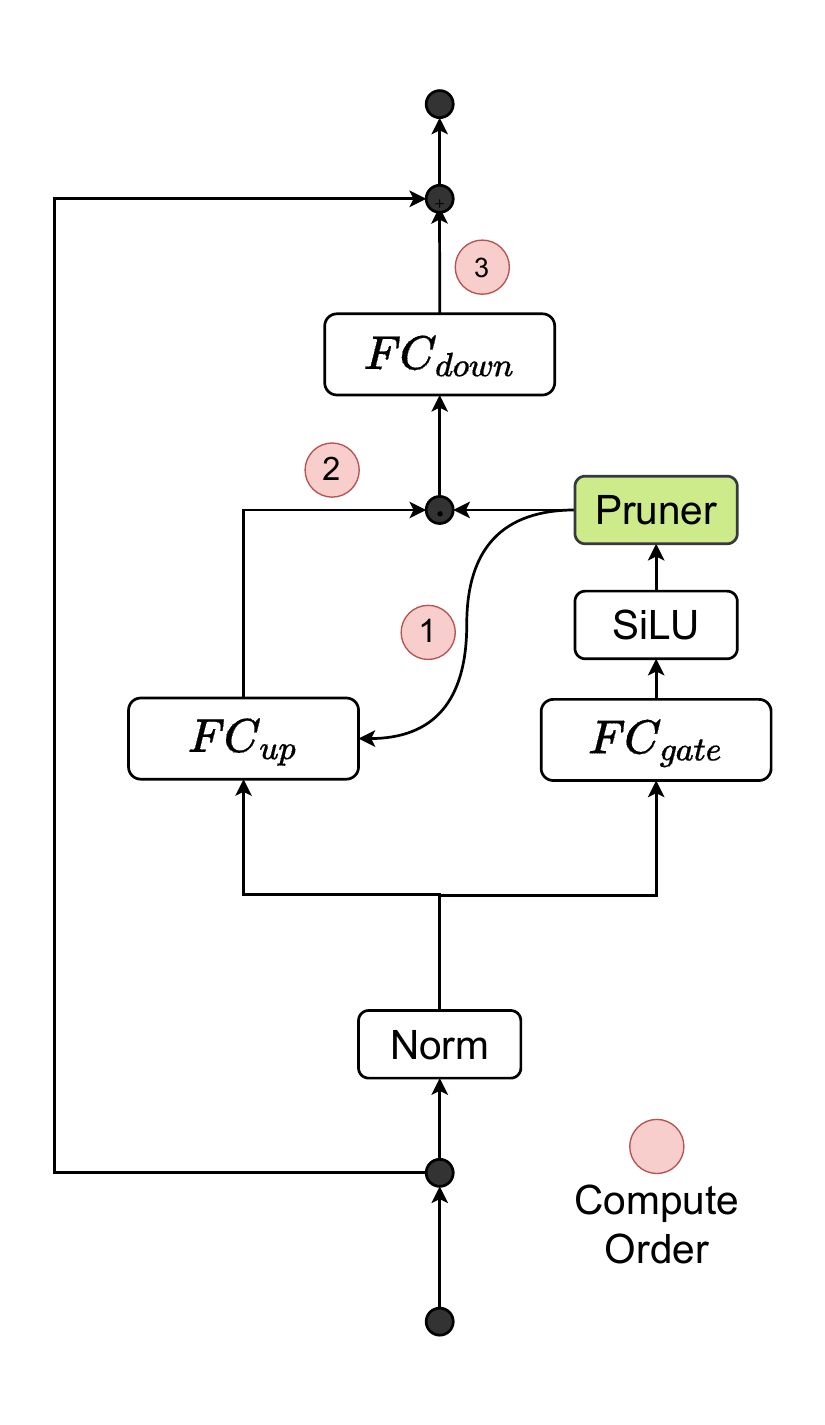}
        \caption{CATS \cite{lee_cats_2024}}
        \label{fig:swiglu_cats}
    \end{subfigure}
    \hfill
    \begin{subfigure}[b]{0.24\textwidth}
        \centering
        \includegraphics[width=\textwidth]{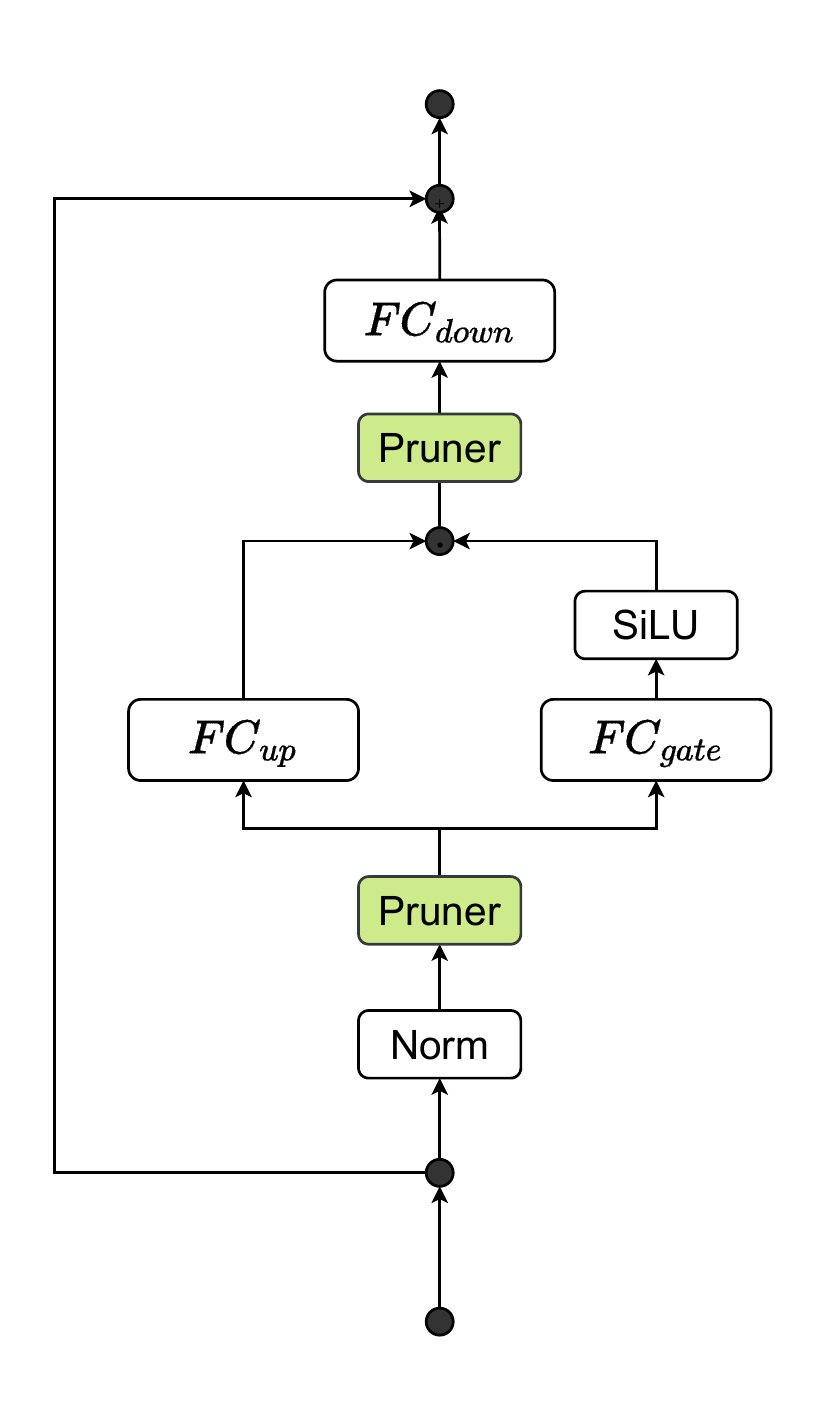}
        \caption{SCAP (Ours)}
        \label{fig:swiglu_scap}
    \end{subfigure}
    \caption{Activation Sparsification across methods on SwiGLU}
    \label{appx:pruner_location}
\end{figure}

Concretely, consider a linear layer in Transformer with input activation \(X\in\mathbb{R}^{N\times IC}\),  weight matrix \(W\in\mathbb{R}^{IC\times OC}\) and bias \(b \in \mathbb{R}^{OC}\), the output activation \(Y \in \mathbb{R}^{N \times OC}\) is given by:
\begin{equation}
Y = XW + b \label{eq:XW}
\end{equation}
where $N$ is batch size, $IC$ and $OC$ are input and output channel dimensions of the weight. The goal is to induce sparsity in input activation \(X\) with a pruning operator. The pruner measures the importance of neurons (activations) on-the-fly and only propagates downstream the neurons with importance above a calibrated threshold. 
\begin{equation}
Pruner(X) = 
    \begin{cases}
    X_{ij}, & \text{if } |X_{ij}| > \tau \\
    0, & \text{otherwise }
    \end{cases}
\text{  , where  }\tau = \text{Quantile}(|X_{calib}|, s)
    \label{eq:pruner}
\end{equation}
Considering the cost of online computation, the importance of each activation (neuron) is calculated using the $L_{1}$ norm $|X_{ij}|$, which is simply the absolute value of the activation. The pruning threshold \texttau is a hyperparameter that correlates with the intensity of pruning and can be determined through calibration. By feeding few-shot of data, the activation tensors of interest can be saved, forming a representative sample for estimating population importance. Subsequently, a quantile function applied to this sample identifies the corresponding importance value for a desired sparsity $s$. For example, setting $\tau = Quantile(|X_{calib}|, 0.3)$ implies that thresholding on $|X|$ with value of \texttau is expected to yield 30\% of sparsity in activation $X$. 

The formulation presented thus far is conceptually similar to CATS\cite{lee_cats_2024}, except that our approach applies pruning to the input activations of the projection layers in Transformers instead of activation function in FFN. The resulting dynamic sparse FC is
\begin{equation}
\text{Sparse FC, }Y' = X'W' + b \label{eq:X'W'}
\end{equation}
In the results, we demonstrate that this approach leads to a more favorable Pareto front and execution scheme. \autoref{appx:target_vs_actual_sparsity} provides empirical evidence showing that the observed sparsity, on average, aligns closely with the target sparsity across a set of 10 downstream tasks. In the subsequent section, we will detail further strategies to enhance the sparsity level of activations with skewed and shifted distributions.


\section{Activation Mode Centering}
\label{sec:activation_mode_centering}
\begin{figure}[htbp]
    \centering

    \begin{minipage}{\textwidth}
        \begin{minipage}[c]{0.05\textwidth}
            \subcaption{}
            \label{fig:layers_of_act_dist(a)}
        \end{minipage}%
        \begin{minipage}[c]{0.95\textwidth}
            \centering
            \includegraphics[width=\textwidth]{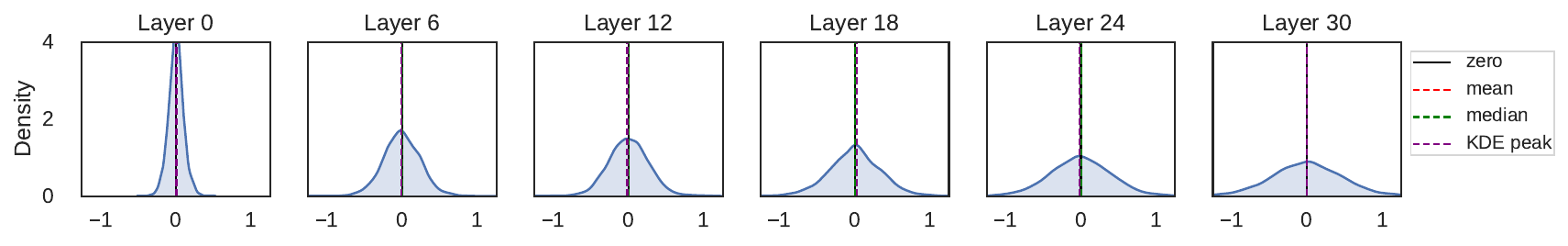}
        \end{minipage}
    \end{minipage}

    \vspace{1em}

    \begin{minipage}{\textwidth}
        \begin{minipage}[c]{0.05\textwidth}
            \subcaption{}
            \label{fig:layers_of_act_dist(b)}
        \end{minipage}%
        \begin{minipage}[c]{0.95\textwidth}
            \centering
            \includegraphics[width=\textwidth]{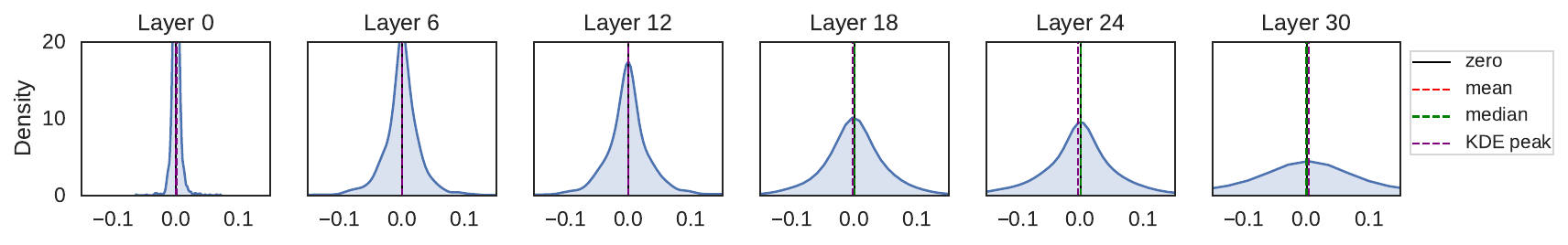}
        \end{minipage}
    \end{minipage}

    \vspace{1em}

    \begin{minipage}{\textwidth}
        \begin{minipage}[c]{0.05\textwidth}
            \subcaption{}
            \label{fig:layers_of_act_dist(c)}
        \end{minipage}%
        \begin{minipage}[c]{0.95\textwidth}
            \centering
            \includegraphics[width=\textwidth]{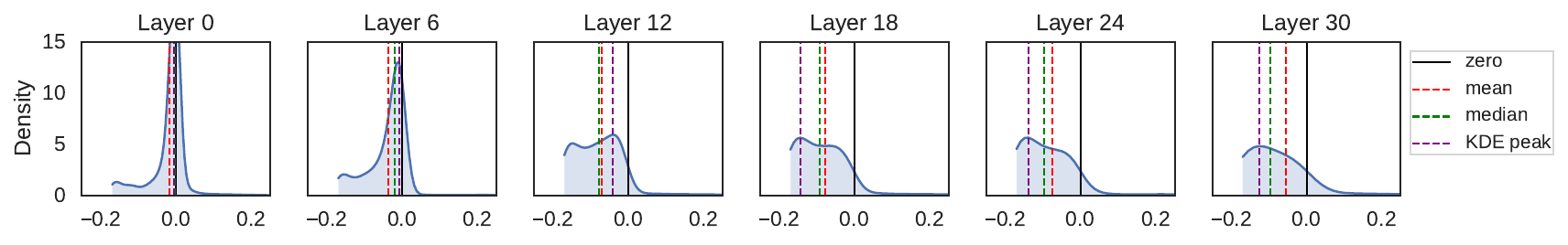}
        \end{minipage}
    \end{minipage}

    \caption{Activation distributions for (a) post-normalization inputs to $FC_{up}$ and $FC_{gate}$; (b) inputs to $FC_{down}$ from gated activations in SwiGLU; and (c) inputs of $FC_{down}$ with a preceding GELU.}
\end{figure}

The prunability of input activations of FC layers depends on the preceding layer. Empirically, we found that not all activations can be sparsified to high levels without compromising task performance. Upon analyzing the distribution of these activations, we identified two primary patterns: one with the mode centered around zero (\autoref{fig:layers_of_act_dist(a)}, \ref{fig:layers_of_act_dist(b)}), and another with the peak away from zero (\autoref{fig:layers_of_act_dist(c)}).

$L_1$-based pruning inherently targets elements within a narrow range around zero, making it particularly effective for zero-centered distributions due to the dense concentration of near-zero values. However, for distributions with a mode away from zero, near-zero elements are less frequent, and achieving higher sparsity requires raising the threshold. This, in turn, introduces non-trivial distortions to the activation representation.

To overcome this limitation, we propose a \textbf{Mode-Centering Calibration} that statistically conditions the targeted activations to center their mode to a near-zero value, which in turn improving prunability with $L_{1}$ thresholding. This calibration, applied prior to activation pruning forms the main ingredient of our proposal, which we name our method as \textbf{Statistical Calibrated Activation Pruning (SCAP)}.


\label{fig:layers_of_act_dist}

\begin{figure}[h]
    \centering
    \begin{subfigure}[t]{0.32\textwidth}
        \centering
        \includegraphics[width=\textwidth]{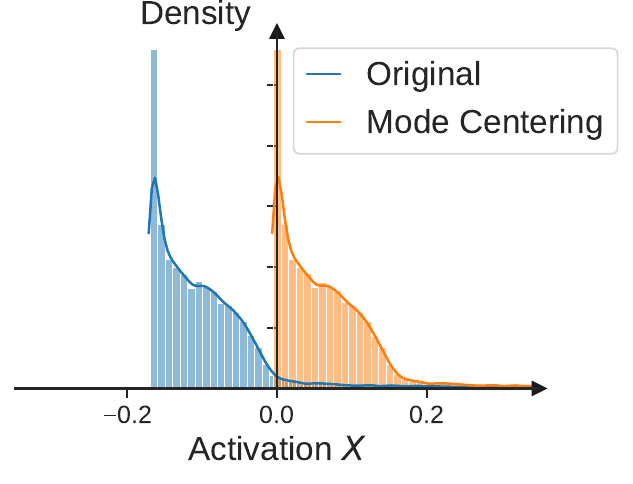}
        \caption{Original and mode-centered distributions of an FC input with a preceding GELU.}
        \label{fig:figure1}
    \end{subfigure}
    \hfill
    \begin{subfigure}[t]{0.32\textwidth}
        \centering
        \includegraphics[width=\textwidth]{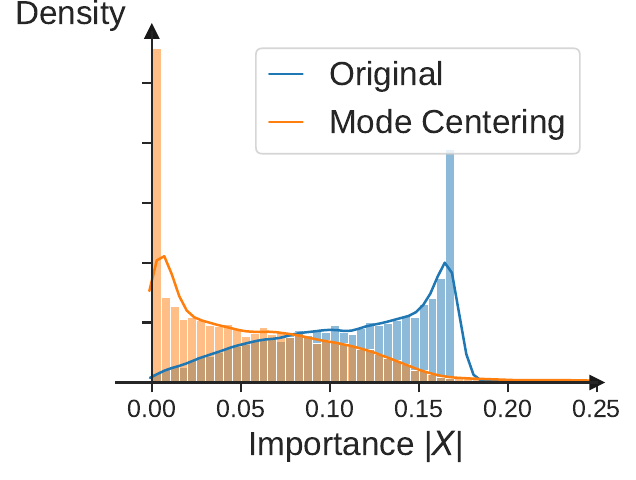}
        \caption{Mode-centered activation results in high density around zero in its $L_1$ importance distribution.}
        \label{fig:figure2}
    \end{subfigure}
    \hfill
    \begin{subfigure}[t]{0.32\textwidth}
        \centering
        \includegraphics[width=\textwidth]{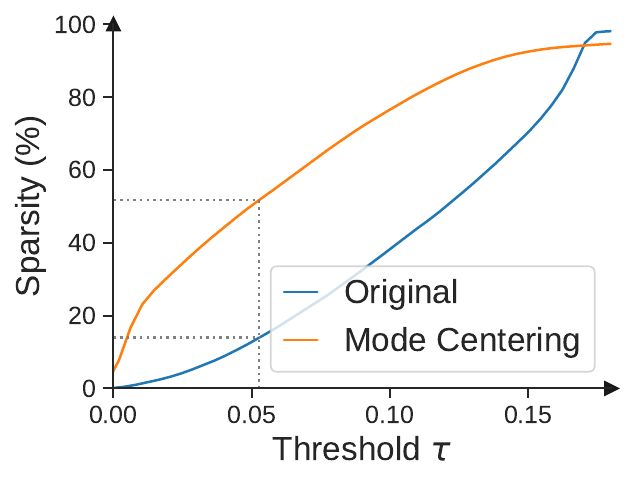}
        \caption{CDF of importance $|X|$ reflects anticipated sparsity. At a threshold of 0.05, mode-centering increases sparsity by about 40 points.}
        \label{fig:figure3}
    \end{subfigure}
    \caption{Effect of Mode-Centering Calibration on Activation Sparsity}
    \label{fig:mode-centering-effect}
\end{figure}

\textbf{The intuition:} \autoref{fig:layers_of_act_dist(c)} shows  GELU output of Falcon-7B at various Transformer blocks, but the peak is at value away from zero, especially those at the deeper Transformer blocks. By shifting the mode (peak) of the activation distribution where the density is highest to zero, more elements with value around zero, further $L_{1}$ thresholding around this region naturally lead to higher sparsity. Refer to \autoref{fig:mode-centering-effect} how mode-centering enhances sparsity.

\textbf{Mode-Centering Calibration:} Consider a FC layer (\autoref{eq:XW}) where its input activation $X$ is the target of sparsification. Let $\eta$ be the mode of $X$,  $\eta$ is introduced to shift the value of elements in $X$, and a compensating $\eta$ is added to maintain functional equivalence. 
\begin{equation}
Y = (X-\eta+\eta)W + b \label{eq:XW_intro_eta}
\end{equation}
Algebraic manipulation results in the following computation should $\eta$ is designed to be determined online and is changing for every input to the model. 
\begin{equation}
Y = (X-\eta)W + \eta W + b \label{eq:etaW_term}
\end{equation}
Dynamic mode $\eta$ incurs higher cost due to just-in-time mode estimation and realization of the compensating $\eta W$. If $\eta$ is static where its value is determined offline during pre-deployment, the compensating term can be fused to the bias $b$ since $W,\eta,b$ are frozen during deployment. The inference overhead is minimal since $\eta$ is a scalar, requiring only broadcast and element-wise subtraction from $X$. 
\begin{equation}
Y = (X-\eta)W + b_{fused} \label{eq:fusedb}
\end{equation}

\textbf{Determination of mode value, $\eta$:} A fast estimation of mode $\eta$ can be obtained through an empirical mean or median of the activations collected over a calibration dataset. As illustrated in \autoref{fig:layers_of_act_dist(c)}, both the mean and median can approximate the actual mode of a distribution. For a more precise estimation of the mode, a probability density estimation algorithm can be employed, many of which are readily available in statistical software. One such example is Kernel Density Estimation (KDE), which is implemented in \texttt{Scipy}. The KDE can be looked up for its the mode value. Since this estimation is typically performed during offline, more involved algorithms are also feasible.

\begin{figure}[htbp]
    \centering
    \includegraphics[width=0.9\textwidth]{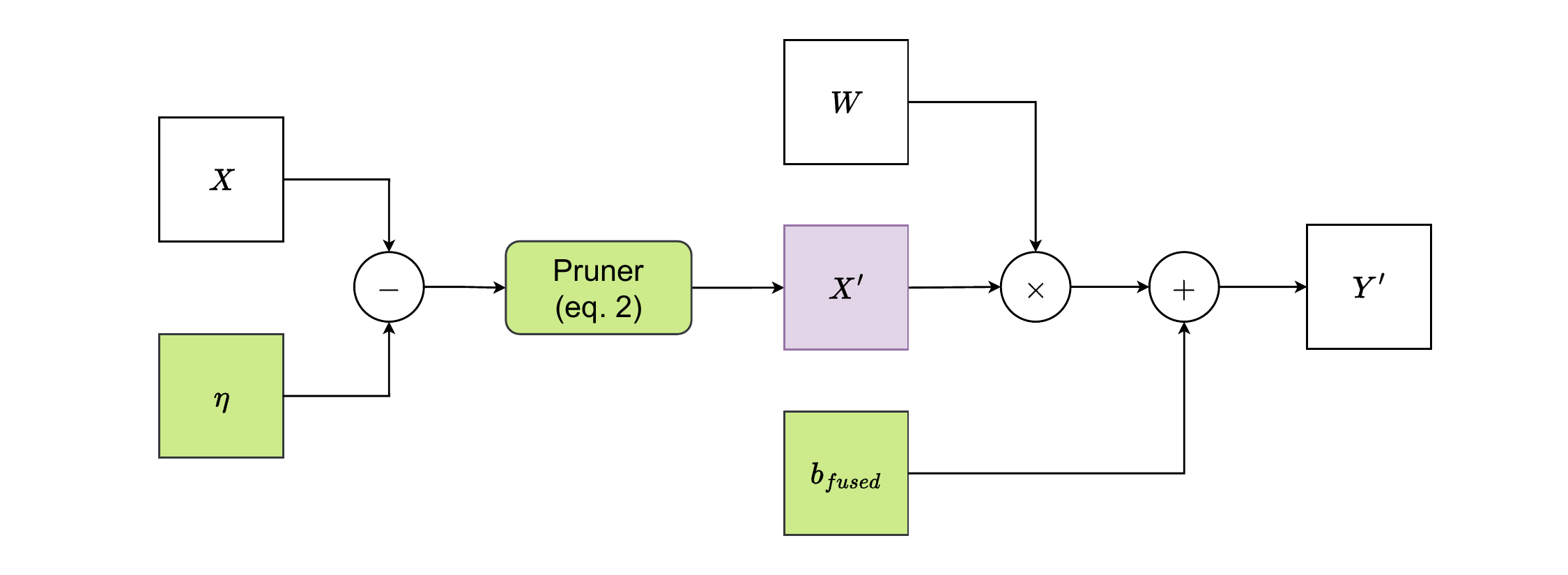}
    \caption{Computational graph of an FC layer with mode-centered and pruned input activation}
    \label{fig:singlefigure}
\end{figure}

We note that the mode and pruning threshold determination can be carried out sequentially within the same calibration process, hence the optimization turnaround remains largely unaffected. Mode-centering, however, has minimal impact when the peak density of the target activation is already near zero.

\newpage
\section{Results and Discussions}
\label{sec:experiments}

\textbf{Implementation:} Our experiments compared SCAP to contemporary activation sparsification methods, including post-training CATS\cite{lee_cats_2024} and TurboSparse\cite{song_turbo_2024}, a state-of-the-art Relufication technique. We closely aligned our setup with these works by focusing on sparsity within the FFN layers, as most of these methods do. Specifically, we targeted two pruning locations: one at the input of the Up projection or the common activation that fans into the Up and Gate projections of the GLU-FFN, and the other at the input of the Down projection.

We grouped activations by these locations across Transformer blocks to prune at a uniform target sparsity, hence requiring two sparsity levels corresponding to the two groups. We swept the two axes in grid, with increments of 10\% (or down to 5\% in some cases) within the 20-80\% range. SCAP used a calibration set of 64 text slices, each consisting of 256 tokens sampled from the C4 dataset\cite{2019t5}, a validation set from WikiText\cite{merity2016pointer}, and downstream tasks aligning with the target comparison method as the test set. We note that mode-centering calibration was only applied to non-GLU FFN, as activations in GLU were observed to be centered. Further details on the experiments discussed below can be found in \autoref{appx:exp_details}.

\subsection{Pareto Efficiency in Tasks vs. Activation Sparsity}
\label{subsec:pareto}

\begin{figure}[h]
    \centering

\includegraphics[width=0.45\textwidth]{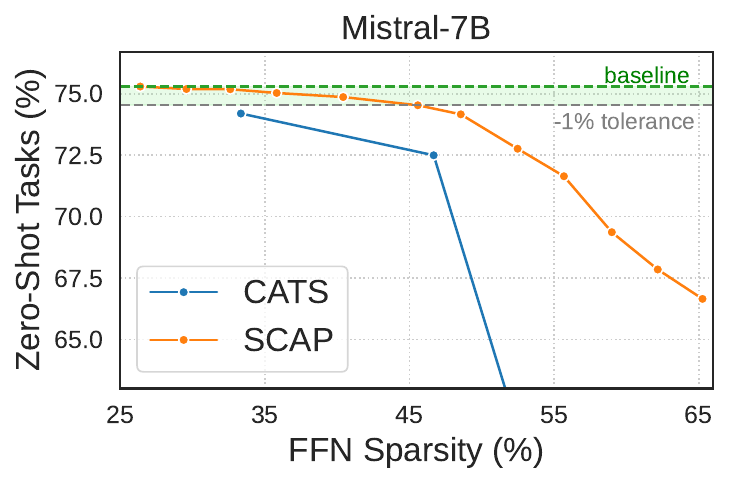}
\hfill
\includegraphics[width=0.43\textwidth]{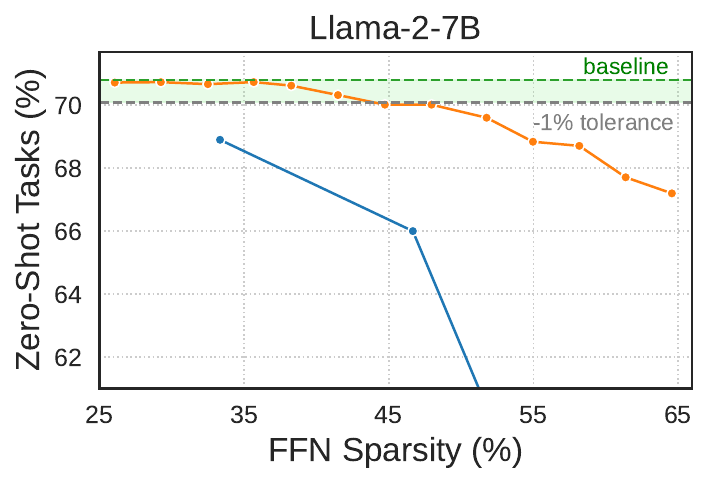} 
     
    \caption{Pareto front of CATS and SCAP (Ours) across LMs, with numbers provided in \autoref{appx:pareto}}
    \label{fig:SCAPvCAT_pareto}
\end{figure}

Compared to CATS across the Mistral-7B-v0.1 and Llama-2-7B models, SCAP consistently maintains accuracy close to baseline zero-shot tasks while achieving higher FFN sparsity. This demonstrates SCAP’s Pareto-efficient trade-off between sparsity and task performance. Although the task trade-off is use-case dependent, SCAP offers multiple viable candidates within the commonly accepted -1\% tolerance in task accuracy, as highlighted in the shaded region of the \autoref{fig:SCAPvCAT_pareto}. 

The sharper decline observed in CATS is attributed to its sole reliance on post-SiLU sparsification, limiting optimization to a single axis and enforcing shared sparse channels between the Up and Down projectors, thereby forgoing alternative sparsity combinations. CATS also overlooks the sparsity opportunities in the Gate projection. In contrast, SCAP applies sparsification more broadly across input activations of FC layers, leading to higher FFN sparsity that effectively utilizes all three FC layers in the SwiGLU FFN. 

A detailed breakdown of sparsity in the accompanying \autoref{tab:pareto_mistral} and \ref{tab:pareto_llama} further illustrates that the Down projector's activations are more prunable than those of the Up projector. This highlights the importance of flexibility in layer-specific sparsification to achieve robust compression. Our preliminary results, obtained through a grid search on two group-level sparsity, validate the trade-off efficiency of this approach. The exploration of a more fine-grained, layer-wise sparsity search is deferred to future studies.

\subsection{Decoding Speedup}
\label{subsec:speedup}

Our kernel implementation is discussed at length in \autoref{appx:kernel} and the latency is confirmed to be comparable to CATS’ which scales proportionally to sparsity. The primary interest of this section is the actual acceleration of decoding stage by activation sparsity. From \autoref{fig:SCAPvCAT_pareto}, we selected a pair of CATS and SCAP-pruned Mistral-7B models that are near-equivalent in task performance, and benchmarked them for 128-token generation with varying input prompt lengths. The results are presented in \autoref{tab:generate_speedup}.

\begin{wraptable}{r}{0.5\textwidth} 

\caption{Inter-Token Latency Speedup (against dense) on Mistral-7B at Iso-Quality.}
\label{tab:generate_speedup}
\centering
\begin{tabular}{ccc}
\toprule
 & \textbf{CATS} & \textbf{SCAP} \\
\midrule
\textbf{FFN Sparsity} & 33.3\% & \textbf{48.5\%} \\
\textbf{Zero-shot Tasks} & 74.2\% & 74.2\% \\
\midrule
\textbf{Prompt Length} & \multicolumn{2}{c}{\textbf{Decoding Speedup}} \\
\midrule
256 & 19.9\% & \textbf{30.4\%} \\
512 & 18.3\% & \textbf{27.9\%} \\
1024 & 16.5\% & \textbf{25.0\%} \\
2048 & 16.4\% & \textbf{25.5\%} \\
\midrule
\textbf{Geomean} & 17.7\% & \textbf{27.1\%} \\
\bottomrule
\end{tabular}
\end{wraptable}

The table reveals that SCAP consistently outperforms CATS in decoding speedup. Underscored by the geometric mean across experimented prompt lengths, SCAP achieves a speedup of 27.1\%, compared to CATS’s 17.7\%. Crucially, SCAP extends CATS's speedup by 1.5$\times$ (27.1\%/17.7\%). This gain stems from the higher FFN sparsity achievable by SCAP while maintaining quality at the same level.

The diminishing speedup with increasing prompt length in both methods is expected, as the growing runtime contribution of attention reduces the impact of sparse GEMV layers. This could be mitigated by combining activation sparsity with efficient attention or KV compression methods.


\subsection{Ablations of Activation Mode Centering}
\label{subsec:ablate_mode_center}

\begin{figure}[h]
    \centering
    \begin{subfigure}[t]{0.46\textwidth}
        \centering
        \includegraphics[width=\textwidth]{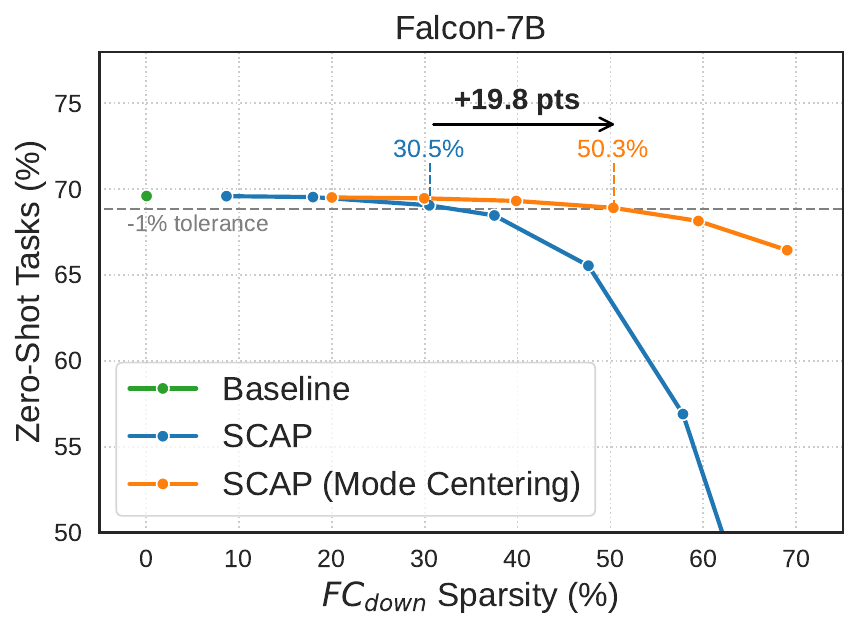}
        \caption{Mode estimated by Median}
        \label{fig:falcon_7b_pareto}
    \end{subfigure}
    \hfill
    \begin{subfigure}[t]{0.46\textwidth}
        \centering
        \includegraphics[width=\textwidth]{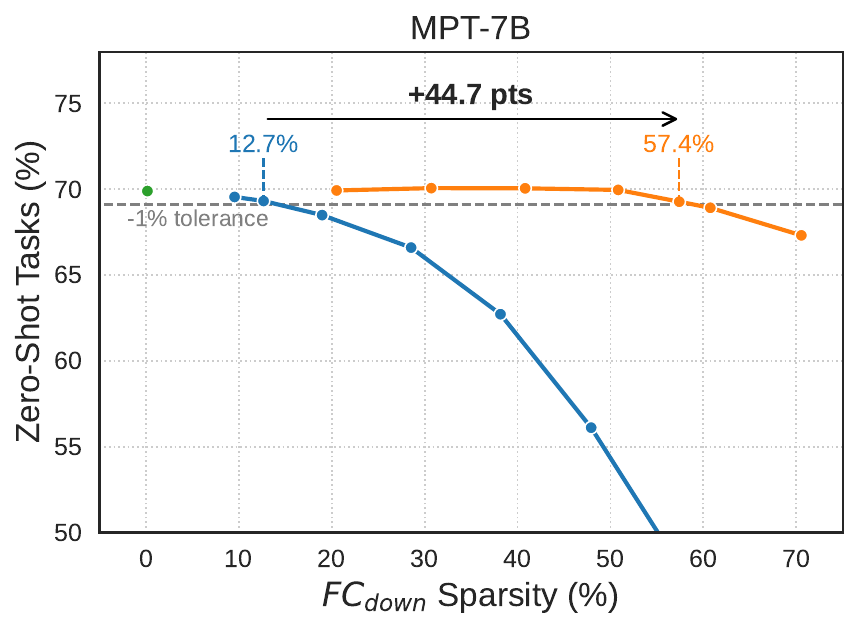} 
        \caption{Mode estimated by \texttt{Scipy KDE}}
        \label{fig:mpt_7b_pareto}
    \end{subfigure}
    \caption{Activation Pruning with and without Mode Centering on input of Down Projection (GELU output) for Falcon-7B and MPT-7B.}
    \label{fig:ablate_mode_centering}
\end{figure}

Non-GLU based LLMs like the Falcon\cite{almazrouei_falcon_2023} and MPT\cite{mosaicml_mpt_2023} families exhibited limited prunability with post-training sparsification methods, particularly in the input activation to the Down projection which originates from the GELU function. As shown in \autoref{fig:ablate_mode_centering}, at a -1\% relative drop in a set of zero-shot tasks, Falcon-7B achieved only 30.5\% sparsity, while MPT-7B struggled even more, with only 12.7\% sparsity. 

Applying SCAP's Mode-Centering technique, where activations are shifted by the estimated mode, allows for significant sparsity through subsequent $L_1$ magnitude thresholding without compromising quality. Notably, for Falcon-7B, the exploitable sparsity in the Down projector increased by 1.6 times, rising from 30.5\% to 50.3\%. MPT-7B showed an even more remarkable improvement, with sparsity jumping by 44.7 points, from 12.7\% to 57.4\%. We further confirm that mode-centering is also applicable to Transformer encoder and vision modalities (see bottom 2 rows of \autoref{tab:many_models}).

\subsection{Comparison to SOTA Relufication}
\label{subsec:scap_v_turbosparse}

More elaboration on TurboSparse and the consideration of SCAP input model can be found in \autoref{appx:details_turbosparse}. From \autoref{tab:scap_v_turbosparse}, TurboSparse \cite{song_turbo_2024} achieved significantly higher FFN sparsity levels compared to SCAP, reaching 82.2\% versus SCAP’s 42.3\%. This was primarily driven by two factors: (1) the retrofitting of two ReLUs, enabling a staggering 91.1\% sparsity in the Down projector, and (2) the use of a predictor that identified and skipped sparse output channels in the Up and Gate FCs. TurboSparse also demonstrated a higher average score across the tasks governed by the OpenLLM leaderboard. This was largely due to its outsized performance on GSM8K (65.7\% vs. 37.9\%), which significantly elevated its overall average. It is important to note that TurboSparse benefited from a set curated pretraining and SFT data, including math-related \cite{paster2023openwebmath, mitra2024orcamath}, which contributed to its efficacy on the Math dataset GSM8K and MMLU.

For the remaining tasks, SCAP-pruned Mistral Instruct models outperformed TurboSparse while achieving 42.3\% FFN sparsity. This was accomplished without the need for hundreds of billions of tokens of uptraining, downstream instruction tuning, predictor training, or the significant demands of datacenter-class GPUs. While TurboSparse's task performance may potentially benefit from further training for greater language modeling, post-training methods like SCAP democratize activation sparsification with its drastic lower computing resource needs, and controllable sparsity-task trade-off as shown in \autoref{subsec:pareto}. On sparsity front, we hypothesize that combining SCAP with parameter-efficient fine-tuning could push further.


\begin{table}[htbp]
\caption{Post-Training vs Training-based Activation Sparsification (SCAP vs TurboSparse)}
\label{tab:scap_v_turbosparse}
\centering
\resizebox{0.75\textwidth}{!}{%
\begin{tabular}{lccc}
\toprule
\textbf{}                                   & \textbf{Mistral-7B} & \textbf{SCAP}        & \textbf{TurboSparse} \\ 
                                            & (Instruct v0.2)     &                     &                      \\ 
\midrule
\textbf{ARC-challenge} (25-shot)            & 63.0                & \textbf{63.0}       & 62.5                 \\ 
\textbf{Hellaswag} (5-shot)                 & 84.8                & \textbf{84.5}       & 82.0                 \\ 
\textbf{MMLU} (5-shot)                      & 60.8                & 60.0       & \textbf{64.0}                 \\ 
\textbf{TruthfulQA} (0-shot)                & 68.2                & \textbf{67.3}       & 52.6                 \\ 
\textbf{WinoGrande} (5-shot)                & 76.6                & \textbf{76.4}       & 74.6                 \\ 
\textbf{GSM8K} (5-shot)                     & 39.0                & 37.9                & \textbf{65.7}        \\
\cmidrule(lr){1-1}\cmidrule(lr){2-2}\cmidrule(lr){3-3}\cmidrule(lr){4-4}
\textbf{OpenLLM Leaderboard Average}        & 65.4                & 64.8                & \textbf{66.9}        \\ 
\midrule
\textbf{Input Sparsity}                     &                     &                     &                      \\ 
Up                                          & -                   & 32.4                & 77.8                 \\ 
Gate                                        & -                   & 32.4                & 77.8                 \\ 
Down                                        & -                   & 62.3                & 91.1                 \\ 
\cmidrule(lr){1-1}\cmidrule(lr){2-2}\cmidrule(lr){3-3}\cmidrule(lr){4-4}
\textbf{FFN}                                & -                   & 42.3                & \textbf{82.2}        \\ 
\midrule
\textbf{Optimization Infrastructure}        &                     & \textbf{1$\times$A100-80GB}  & 64$\times$A800-80GB  \\ 
\bottomrule
\end{tabular}
}
\end{table}

\section{Conclusions}
\label{sec:conclusion}
In this work, we developed Statistical-Conditioned Activation Pruning (SCAP), a post-training method that effectively induces activation sparsity in LLMs without the need for sparsification training. By focusing on input activations to FC layers, SCAP generalizes both pruning and implementation within Transformer, achieving a greater trade-off between computational efficiency and task performance compared to prior methods. The introduction of Mode-Centering pre-calibration addresses the limited post-training sparsity in activations with non-zero mode, leading to substantial increases in their sparsity. Our experimental results validate these findings. We further present \autoref{tab:many_models} detailing SCAP’s applicability across a range of Transformer models, including pre-quantized ones and the emerging MoE and Mamba2, all within a -1\% relative task accuracy from their baselines while attaining sufficient sparsity at targeted activations. This work highlights the potential of post-training method for activation sparsity, while laying aspects for future explorations in \autoref{appx:batch_inference_challenge}.

\begin{table}[H]
\caption{More SCAP-pruned models. * indicates mode-centering. Details are at Appendix \ref{appx:details_conclusion}.}
\label{tab:many_models}

\centering
\resizebox{0.9\textwidth}{!}{%
\begin{tabular}{l c c c}
\toprule
\textbf{Model}                     & \textbf{Task Relative (\%)} & \textbf{Sparsity} & \textbf{Target Sparsity, $s_{layer}$ (\%)}                         \\ 
\midrule
\href{https://huggingface.co/meta-llama/Llama-2-7b-hf}{Llama-2-7B}                         & -0.8\%                           & 42\%                   & $s_\text{up/gate}$: 35, $s_\text{down}$: 55                 \\ 
\href{https://huggingface.co/meta-llama/Llama-2-13b-hf}{Llama-2-13B}                        & -0.7\%                           & 47\%                   & $s_\text{up/gate}$: 40, $s_\text{down}$: 60                 \\ 
\href{https://huggingface.co/meta-llama/Llama-2-70b-hf}{Llama-2-70B}                        & -0.9\%                           & 50\%                   & $s_\text{up/gate}$: 45, $s_\text{down}$: 60                 \\ 
\href{https://huggingface.co/TheBloke/Llama-2-70B-Chat-AWQ}{Llama-2-70B (4-bit)}             & -0.9\%                           & 43\%                   & $s_\text{up/gate}$: 35, $s_\text{down}$: 60                 \\ 
\midrule
\href{https://huggingface.co/meta-llama/Meta-Llama-3.1-8B-Instruct}{Llama-3.1-8B-Instruct}        & -0.7\%                           & 43\%                   & $s_\text{up/gate}$: 40, $s_\text{down}$: 50                 \\ 
Llama-3.1-8B-Instruct (8-bit) & -0.8\%                           & 43\%                   & $s_\text{up/gate}$: 40, $s_\text{down}$: 50                 \\ 
\midrule

\href{https://huggingface.co/tiiuae/falcon-7b}{Falcon-7B*} & -0.8\%     & 38\%                   & $s_\text{up}$: 30, $s_\text{down}$: 45                 \\ 
\midrule

\href{https://huggingface.co/mosaicml/mpt-7b}{MPT-7B*} & -0.6\%     & 43\%                   & $s_\text{up}$: 40, $s_\text{down}$: 45                 \\ 
\midrule

\href{https://huggingface.co/mistralai/Mixtral-8x7B-Instruct-v0.1}{Mixtral-8x7B-Instruct-v0.1}         & -0.8\%                           & 43\%                   & $s_\text{up/gate}$: 40, $s_\text{down}$: 50                 \\ 
\href{https://huggingface.co/casperhansen/mixtral-instruct-awq}{Mixtral-8x7B-Instruct-v0.1 (4-bit)} & -1.0\%                           & 43\%                   & $s_\text{up/gate}$: 40, $s_\text{down}$: 50                 \\ 
\midrule

\href{https://huggingface.co/state-spaces/mamba2-2.7b}{Mamba2-2.7B}                        & -0.8\%                           & 40\%                   & $s_\text{in}$: 30, $s_\text{out}$: 60                       \\ 
\midrule
\href{https://huggingface.co/timm/deit_base_patch16_224.fb_in1k}{DeiT-base*}                          & -0.9\%                           & 51\%                   & $s_\text{qkv}$: 40, $s_\text{o}$: 50, $s_\text{up}$: 55, $s_\text{down}$: 55    \\ 
\href{https://huggingface.co/timm/deit3_large_patch16_384.fb_in1k}{DeiT3-large*}                        & -0.9\%                           & 59\%                   & $s_\text{qkv}$: 60, $s_\text{o}$: 30, $s_\text{up}$: 60, $s_\text{down}$: 65    \\ 
\bottomrule
\end{tabular}%
}
\end{table}

\newpage
\begin{ack}
The authors would like to thank Intel Labs and the OpenVINO team for their valuable discussions and support throughout this work. We are especially grateful to Tingqian Li, Cheng Luo, Xian Fu Wong, Gopi Krishna Jha, Nikita Savelyev, and Alexander Kozlov for their collaborative efforts and contributions, which enriched the development of this work.
\end{ack}

\medskip  

\bibliographystyle{plain} 
\bibliography{references}

\begin{thebibliography}{10}

\bibitem{akhauri_shadowllm_2024}
Yash Akhauri, Ahmed~F. AbouElhamayed, Jordan Dotzel, Zhiru Zhang, Alexander~M. Rush, Safeen Huda, and Mohamed~S. Abdelfattah.
\newblock {ShadowLLM}: {Predictor}-based {Contextual} {Sparsity} for {Large} {Language} {Models}, June 2024.
\newblock arXiv:2406.16635 [cs].

\bibitem{alizadeh_llm_2024}
Keivan Alizadeh, Iman Mirzadeh, Dmitry Belenko, Karen Khatamifard, Minsik Cho, Carlo~C. Del~Mundo, Mohammad Rastegari, and Mehrdad Farajtabar.
\newblock {LLM} in a flash: {Efficient} {Large} {Language} {Model} {Inference} with {Limited} {Memory}, July 2024.
\newblock arXiv:2312.11514 [cs].

\bibitem{almazrouei_falcon_2023}
Ebtesam Almazrouei, Hamza Alobeidli, Abdulaziz Alshamsi, Alessandro Cappelli, Ruxandra Cojocaru, Mérouane Debbah, Étienne Goffinet, Daniel Hesslow, Julien Launay, Quentin Malartic, Daniele Mazzotta, Badreddine Noune, Baptiste Pannier, and Guilherme Penedo.
\newblock The {Falcon} {Series} of {Open} {Language} {Models}, November 2023.
\newblock arXiv:2311.16867 [cs].

\bibitem{chen2021evaluatinglargelanguagemodels}
Mark Chen, Jerry Tworek, Heewoo Jun, Qiming Yuan, Henrique~Ponde de~Oliveira~Pinto, Jared Kaplan, Harri Edwards, Yuri Burda, Nicholas Joseph, Greg Brockman, Alex Ray, Raul Puri, Gretchen Krueger, Michael Petrov, Heidy Khlaaf, Girish Sastry, Pamela Mishkin, Brooke Chan, Scott Gray, Nick Ryder, Mikhail Pavlov, Alethea Power, Lukasz Kaiser, Mohammad Bavarian, Clemens Winter, Philippe Tillet, Felipe~Petroski Such, Dave Cummings, Matthias Plappert, Fotios Chantzis, Elizabeth Barnes, Ariel Herbert-Voss, William~Hebgen Guss, Alex Nichol, Alex Paino, Nikolas Tezak, Jie Tang, Igor Babuschkin, Suchir Balaji, Shantanu Jain, William Saunders, Christopher Hesse, Andrew~N. Carr, Jan Leike, Josh Achiam, Vedant Misra, Evan Morikawa, Alec Radford, Matthew Knight, Miles Brundage, Mira Murati, Katie Mayer, Peter Welinder, Bob McGrew, Dario Amodei, Sam McCandlish, Ilya Sutskever, and Wojciech Zaremba.
\newblock Evaluating large language models trained on code, 2021.

\bibitem{chowdhery_palm_2022}
Aakanksha Chowdhery, Sharan Narang, Jacob Devlin, Maarten Bosma, Gaurav Mishra, Adam Roberts, Paul Barham, Hyung~Won Chung, Charles Sutton, Sebastian Gehrmann, Parker Schuh, Kensen Shi, Sasha Tsvyashchenko, Joshua Maynez, Abhishek Rao, Parker Barnes, Yi~Tay, Noam Shazeer, Vinodkumar Prabhakaran, Emily Reif, Nan Du, Ben Hutchinson, Reiner Pope, James Bradbury, Jacob Austin, Michael Isard, Guy Gur-Ari, Pengcheng Yin, Toju Duke, Anselm Levskaya, Sanjay Ghemawat, Sunipa Dev, Henryk Michalewski, Xavier Garcia, Vedant Misra, Kevin Robinson, Liam Fedus, Denny Zhou, Daphne Ippolito, David Luan, Hyeontaek Lim, Barret Zoph, Alexander Spiridonov, Ryan Sepassi, David Dohan, Shivani Agrawal, Mark Omernick, Andrew~M. Dai, Thanumalayan~Sankaranarayana Pillai, Marie Pellat, Aitor Lewkowycz, Erica Moreira, Rewon Child, Oleksandr Polozov, Katherine Lee, Zongwei Zhou, Xuezhi Wang, Brennan Saeta, Mark Diaz, Orhan Firat, Michele Catasta, Jason Wei, Kathy Meier-Hellstern, Douglas Eck, Jeff Dean, Slav Petrov, and Noah Fiedel.
\newblock {PaLM}: {Scaling} {Language} {Modeling} with {Pathways}, October 2022.
\newblock arXiv:2204.02311 [cs].

\bibitem{deng_imagenet_2009}
Jia Deng, Wei Dong, Richard Socher, Li-Jia Li, Kai Li, and Li~Fei-Fei.
\newblock Imagenet: A large-scale hierarchical image database.
\newblock In {\em 2009 IEEE Conference on Computer Vision and Pattern Recognition}, pages 248--255, 2009.

\bibitem{lighteval}
Clémentine Fourrier, Nathan Habib, Thomas Wolf, and Lewis Tunstall.
\newblock Lighteval: A lightweight framework for llm evaluation, 2023.

\bibitem{frantar_gptq_2023}
Elias Frantar, Saleh Ashkboos, Torsten Hoefler, and Dan Alistarh.
\newblock {GPTQ}: {Accurate} {Post}-{Training} {Quantization} for {Generative} {Pre}-trained {Transformers}, March 2023.
\newblock arXiv:2210.17323 [cs].

\bibitem{eval-harness}
Leo Gao, Jonathan Tow, Baber Abbasi, Stella Biderman, Sid Black, Anthony DiPofi, Charles Foster, Laurence Golding, Jeffrey Hsu, Alain Le~Noac'h, Haonan Li, Kyle McDonell, Niklas Muennighoff, Chris Ociepa, Jason Phang, Laria Reynolds, Hailey Schoelkopf, Aviya Skowron, Lintang Sutawika, Eric Tang, Anish Thite, Ben Wang, Kevin Wang, and Andy Zou.
\newblock A framework for few-shot language model evaluation, 12 2023.

\bibitem{han_learning_2015}
Song Han, Jeff Pool, John Tran, and William Dally.
\newblock Learning both {Weights} and {Connections} for {Efficient} {Neural} {Network}.
\newblock In {\em Advances in {Neural} {Information} {Processing} {Systems}}, volume~28. Curran Associates, Inc., 2015.

\bibitem{kwon2023efficient}
Woosuk Kwon, Zhuohan Li, Siyuan Zhuang, Ying Sheng, Lianmin Zheng, Cody~Hao Yu, Joseph~E. Gonzalez, Hao Zhang, and Ion Stoica.
\newblock Efficient memory management for large language model serving with pagedattention.
\newblock In {\em Proceedings of the ACM SIGOPS 29th Symposium on Operating Systems Principles}, 2023.

\bibitem{lee_cats_2024}
Je-Yong Lee, Donghyun Lee, Genghan Zhang, Mo~Tiwari, and Azalia Mirhoseini.
\newblock {CATS}: {Contextually}-{Aware} {Thresholding} for {Sparsity} in {Large} {Language} {Models}, April 2024.
\newblock arXiv:2404.08763 [cs].

\bibitem{li_lazy_2023}
Zonglin Li, Chong You, Srinadh Bhojanapalli, Daliang Li, Ankit~Singh Rawat, Sashank~J Reddi, Ke~Ye, Felix Chern, Felix Yu, Ruiqi Guo, and Sanjiv Kumar.
\newblock {The} {Lazy} {Neuron} {Phenomenon}: {On} {Emergence} {Of} {Activation} {Sparsity} {In} {Transformers}.
\newblock 2023.

\bibitem{lin_awq_2024}
Ji~Lin, Jiaming Tang, Haotian Tang, Shang Yang, Wei-Ming Chen, Wei-Chen Wang, Guangxuan Xiao, Xingyu Dang, Chuang Gan, and Song Han.
\newblock {AWQ}: {Activation}-aware {Weight} {Quantization} for {LLM} {Compression} and {Acceleration}, July 2024.
\newblock arXiv:2306.00978 [cs].

\bibitem{liu_deja_2023}
Zichang Liu, Jue Wang, Tri Dao, Tianyi Zhou, Binhang Yuan, Zhao Song, Anshumali Shrivastava, Ce~Zhang, Yuandong Tian, Christopher Ré, and Beidi Chen.
\newblock Deja {Vu}: contextual sparsity for efficient {LLMs} at inference time.
\newblock In {\em Proceedings of the 40th {International} {Conference} on {Machine} {Learning}}, volume 202 of {\em {ICML}'23}, pages 22137--22176, Honolulu, Hawaii, USA, July 2023. JMLR.org.

\bibitem{merity2016pointer}
Stephen Merity, Caiming Xiong, James Bradbury, and Richard Socher.
\newblock Pointer sentinel mixture models, 2016.

\bibitem{mirzadeh_relu_2023}
Seyed~Iman Mirzadeh, Keivan Alizadeh-Vahid, Sachin Mehta, Carlo C.~del Mundo, Oncel Tuzel, Golnoosh Samei, Mohammad Rastegari, and Mehrdad Farajtabar.
\newblock {ReLU} {Strikes} {Back}: {Exploiting} {Activation} {Sparsity} in {Large} {Language} {Models}.
\newblock October 2023.

\bibitem{mitra2024orcamath}
Arindam Mitra, Hamed Khanpour, Corby Rosset, and Ahmed Awadallah.
\newblock Orca-math: Unlocking the potential of slms in grade school math, 2024.

\bibitem{paster2023openwebmath}
Keiran Paster, Marco~Dos Santos, Zhangir Azerbayev, and Jimmy Ba.
\newblock Openwebmath: An open dataset of high-quality mathematical web text, 2023.

\bibitem{pope_efficiently_2023}
Reiner Pope, Sholto Douglas, Aakanksha Chowdhery, Jacob Devlin, James Bradbury, Jonathan Heek, Kefan Xiao, Shivani Agrawal, and Jeff Dean.
\newblock Efficiently {Scaling} {Transformer} {Inference}.
\newblock {\em Proceedings of Machine Learning and Systems}, 5:606--624, March 2023.

\bibitem{2019t5}
Colin Raffel, Noam Shazeer, Adam Roberts, Katherine Lee, Sharan Narang, Michael Matena, Yanqi Zhou, Wei Li, and Peter~J. Liu.
\newblock Exploring the limits of transfer learning with a unified text-to-text transformer.
\newblock {\em arXiv e-prints}, 2019.

\bibitem{mosaicml_mpt_2023}
Mosaic~AI Research.
\newblock Introducing {MPT}-{7B}: {A} {New} {Standard} for {Open}-{Source}, {Commercially} {Usable} {LLMs} {\textbar} {Databricks} {Blog}, May 2023.

\bibitem{shazeer_glu_2020}
Noam Shazeer.
\newblock {GLU} {Variants} {Improve} {Transformer}, February 2020.
\newblock arXiv:2002.05202 [cs, stat].

\bibitem{song_prosparse_2024}
Chenyang Song, Xu~Han, Zhengyan Zhang, Shengding Hu, Xiyu Shi, Kuai Li, Chen Chen, Zhiyuan Liu, Guangli Li, Tao Yang, and Maosong Sun.
\newblock {ProSparse}: {Introducing} and {Enhancing} {Intrinsic} {Activation} {Sparsity} within {Large} {Language} {Models}, July 2024.
\newblock arXiv:2402.13516 [cs].

\bibitem{song_powerinfer_2023}
Yixin Song, Zeyu Mi, Haotong Xie, and Haibo Chen.
\newblock {PowerInfer}: {Fast} {Large} {Language} {Model} {Serving} with a {Consumer}-grade {GPU}, December 2023.
\newblock arXiv:2312.12456 [cs].

\bibitem{song_turbo_2024}
Yixin Song, Haotong Xie, Zhengyan Zhang, Bo~Wen, Li~Ma, Zeyu Mi, and Haibo Chen.
\newblock Turbo {Sparse}: {Achieving} {LLM} {SOTA} {Performance} with {Minimal} {Activated} {Parameters}, June 2024.
\newblock arXiv:2406.05955 [cs].

\bibitem{sun_simple_2024}
Mingjie Sun, Zhuang Liu, Anna Bair, and J.~Zico Kolter.
\newblock A {Simple} and {Effective} {Pruning} {Approach} for {Large} {Language} {Models}, May 2024.
\newblock arXiv:2306.11695 [cs].

\bibitem{hugo_deit3_2022}
Hugo Touvron, Matthieu Cord, and Herv\'{e} J\'{e}gou.
\newblock Deit iii: Revenge of the vit.
\newblock In {\em Computer Vision – ECCV 2022: 17th European Conference, Tel Aviv, Israel, October 23–27, 2022, Proceedings, Part XXIV}, page 516–533, Berlin, Heidelberg, 2022. Springer-Verlag.

\bibitem{wang_gptj}
Ben Wang and Aran Komatsuzaki.
\newblock Gpt-j-6b: A 6 billion parameter autoregressive language model, 2021.

\bibitem{Wolf2019TransformersSN}
Thomas Wolf, Lysandre Debut, Victor Sanh, Julien Chaumond, Clement Delangue, Anthony Moi, Pierric Cistac, Tim Rault, R{\'e}mi Louf, Morgan Funtowicz, Joe Davison, Sam Shleifer, Patrick von Platen, Clara Ma, Yacine Jernite, Julien Plu, Canwen Xu, Teven~Le Scao, Sylvain Gugger, Mariama Drame, Quentin Lhoest, and Alexander~M. Rush.
\newblock Transformers: State-of-the-art natural language processing.
\newblock In {\em Conference on Empirical Methods in Natural Language Processing}, 2019.

\bibitem{yu_orca_2022}
Gyeong-In Yu, Joo~Seong Jeong, Geon-Woo Kim, Soojeong Kim, and Byung-Gon Chun.
\newblock Orca: A distributed serving system for {Transformer-Based} generative models.
\newblock In {\em 16th USENIX Symposium on Operating Systems Design and Implementation (OSDI 22)}, pages 521--538, Carlsbad, CA, July 2022. USENIX Association.

\bibitem{zhang_relu2_2024}
Zhengyan Zhang, Yixin Song, Guanghui Yu, Xu~Han, Yankai Lin, Chaojun Xiao, Chenyang Song, Zhiyuan Liu, Zeyu Mi, and Maosong Sun.
\newblock {ReLU$^{2}$} {Wins}: {Discovering} {Efficient} {Activation} {Functions} for {Sparse} {LLMs}, February 2024.
\newblock arXiv:2402.03804 [cs].

\end{thebibliography}

\newpage
\appendix
\section{Target vs Actual Activation Sparsity}
\label{appx:target_vs_actual_sparsity}
\begin{figure}[!htb]
    \centering
    \includegraphics[width=1\linewidth]{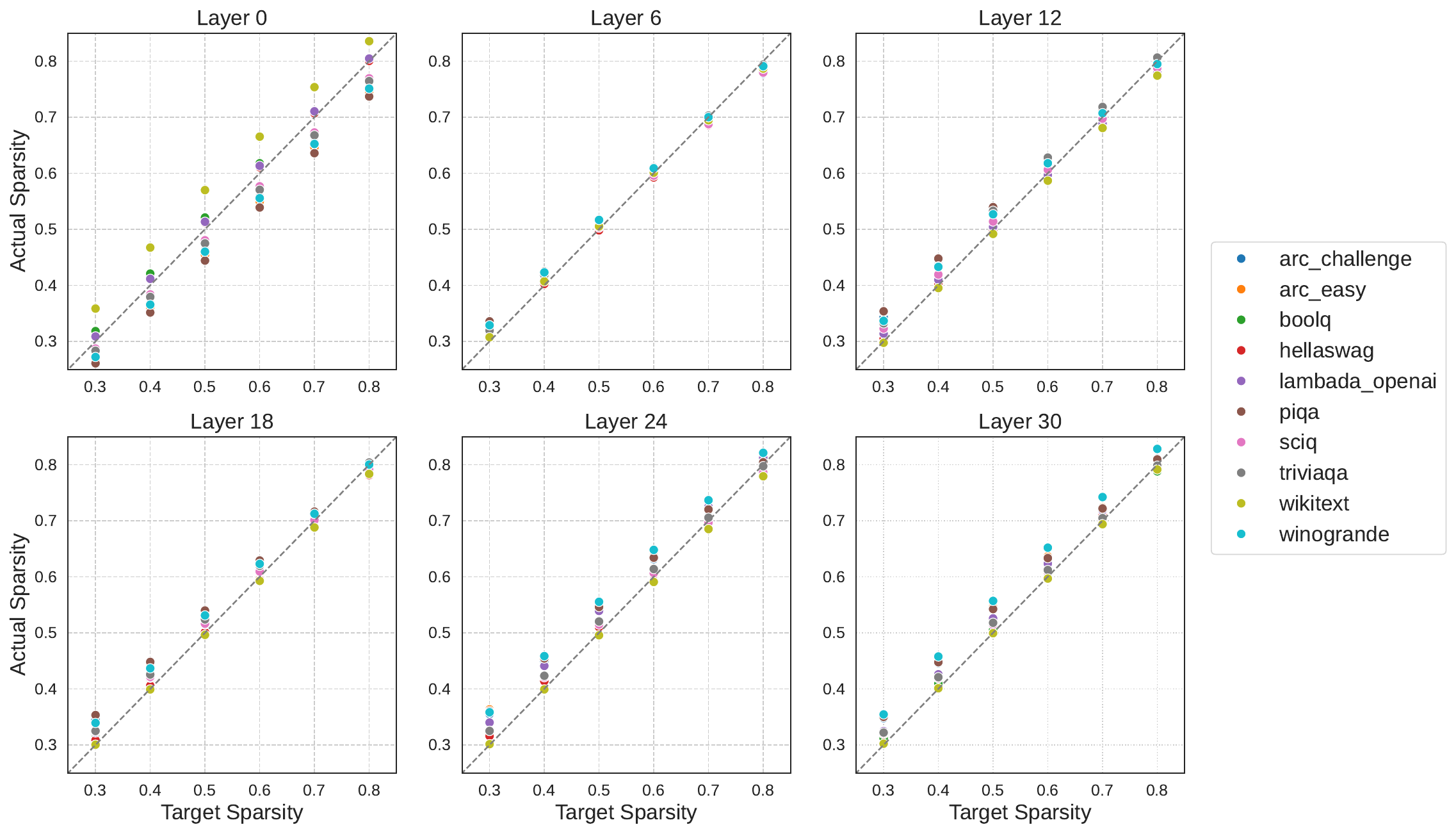}
    \captionof{figure}{Target vs. Actual input Sparsity of $FC_{down}$ across various Transformer Layers in Llama. Each point represents for a different task, with its empirical sparsity closely aligning with diagonal line, indicating a strong correlation with the target. This illustrates the effectiveness of the $L_1$ thresholding based pruning mechanism in maintaining consistent dynamic sparsity across layers for the majority of input prompts.}
    \label{fig:target_v_actual_sparsity}
\end{figure}





\section{Kernel Implementations}
\label{appx:kernel}

\begin{wrapfigure}{l}{0.62\textwidth} 
\vspace{-1em}
\begin{minipage}{0.62\textwidth}
\begin{algorithm}[H]
\caption{CATS' SwiGLU}
\label{alg:cats_swiglu}
\begin{algorithmic}[1]
\State \textbf{Input:} $\tau_{silu}$, $x$, $W_{gate}$, $W_{down}$, $W_{up}$ \as
\State $v \gets silu(xW_{gate})$ \as
\State $\texttt{Mask} \gets 1$ if $|v| \geq \tau_{silu}$ else $0$ \as
\State $x_1 \gets ( x W_{up} [\texttt{Mask}] * v [\texttt{Mask}] )$ \as
\State $y \gets x_{1}W_{down} [\texttt{Mask}]$ \as
\State \textbf{output} $y$
\end{algorithmic}
\end{algorithm}


\begin{algorithm}[H]
\caption{SCAP's SwiGLU (Ours)}
\label{alg:scap_swiglu}
\begin{algorithmic}[1]
\Procedure{scap\_fc}{$\tau, x, W$} \as  
    \State $\texttt{Mask} \gets 1$ if $|x| \geq \tau$ else $0$ \as
    \State $y \gets x_{\eta}W$ \as
    \State \textbf{return} $y$ \as
\EndProcedure

\Statex
\State \textbf{Input:} $\tau_{x},\tau_{gated}$, $x$, $W_{gate}$, $W_{up}$, $W_{down}$ \as
\State $z_{gated} \gets $ \as
\Statex \hspace{\algorithmicindent} $\Call{scap\_fc}{\tau_{x}, x, W_{up}} * silu(\Call{scap\_fc}{ \tau_{x}, x, W_{gate}})$ \as
\State $y \gets \Call{scap\_fc}{\tau_{gated}, x, W_{down}}$ \as
\State \textbf{output} $y$
\end{algorithmic}
\end{algorithm}
\end{minipage}
\end{wrapfigure}

\begin{figure}[H]
    \centering
    \begin{minipage}{0.45\textwidth}
        \begin{algorithm}[H]
        \caption{SCAP FC with Mode Centering}
        \label{alg:demode_scap_fc}
        \begin{algorithmic}[1]
        \Procedure{scap\_fc\_demode}{}   
            \Statex \hspace{\algorithmicindent} $(\tau, \eta, x, W, b_{fused})$ \as
            \State $x_{\eta} \gets x - \eta$ \as       
            \State $\texttt{Mask} \gets 1$ if $|x_{\eta}| \geq \tau$ else $0$ \as
            \State $y \gets x_{\eta}W + b_{fused}$ \as
            \State \textbf{return} $y$ \as
        \EndProcedure
        \end{algorithmic}
        \end{algorithm}
    \end{minipage}
    \hfill
    \begin{minipage}{0.45\textwidth}
        \centering
        \includegraphics[width=\textwidth]{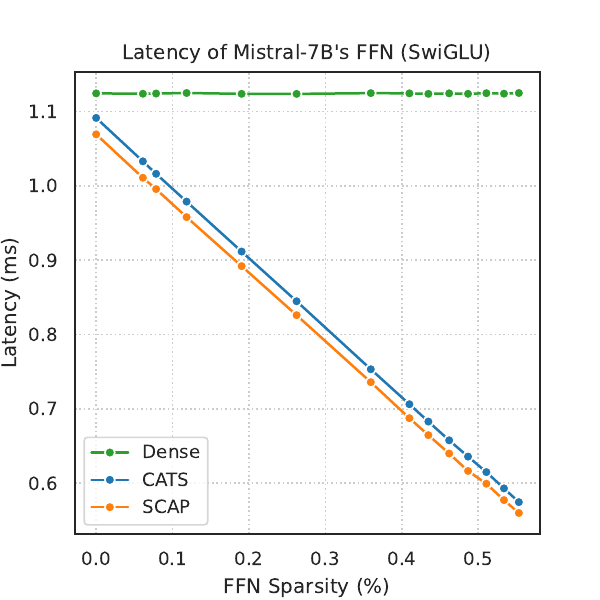}
        \caption{SwiGLU Latency Scaling w.r.t FFN Sparsity across implementations}
        \label{fig:swiglu_latency_sparsity_sweep}
    \end{minipage}
\end{figure}

SCAP proposes a generic sparse-by-input activation across targeted FC layers, entailing a single kernel implementation. This implementation, referred to as \texttt{SCAP\_FC} in Alg. \ref{alg:scap_swiglu}, provides a bias-free version of Eq. 2 and 3. If mode-centering is applied, the corresponding mode shifting and realization of Eq. 6 is detailed in Alg. \ref{alg:demode_scap_fc}. These procedures are separated for brevity but can be merged.

Alg. \ref{alg:cats_swiglu} and \ref{alg:scap_swiglu} compare the SwiGLU kernel between CATS and SCAP. CATS requires a rigid computation order of gate projection path first and a sparse mask coupled for the Up and Down weights. In contrast, SCAP demonstrates the reusability of \texttt{SCAP\_FC}, which can also be applicable to FCs in attention block if their inputs are sparsified. 

We implemented SCAP kernels by adapting the official CATS codes and performed latency benchmarks with a sweep of FFN sparsity on a single NVIDIA L40S GPU (consistent with CATS' experiments). Since the FFN is composed of three equal-sized layers, FFN sparsity $= 2/3\times s_{silu} = 2/3\times s_{x}+1/3\times s_{gated}$. \autoref{fig:swiglu_latency_sparsity_sweep} shows that SCAP’s SwiGLU performs similarly to, or marginally faster than, CATS’s, which is expected given that the memory and compute savings are proportionate in both cases. The plot is clipped at about 55\% of FFN sparsity due to CATS reaching over 80\% sparsity in the Up and Down projectors. We anticipate proportional improvements if SCAP prunes to a lower level of FFN sparsity.

The main advantage of SCAP is its flexibility in decoupling the sparsity between the Gate, Up, and Down inputs, resulting in a favorable Pareto trade-off in task performance, as experimented in \autoref{subsec:pareto}. This means achieving higher sparsity at a given task tolerance, leading to overall higher acceleration, as confirmed in \autoref{tab:generate_speedup}.

\section{Acceleration Challenges of Batched Sparse Activation}
\label{appx:batch_inference_challenge}
Inference acceleration through activation sparsity has primarily focused on the token-to-token decoding phase of language models, operating under the assumption that a single dynamic sparse activation vector can trigger a structured sparse weight pattern, alleviating memory bottlenecks. While many studies demonstrate significant acceleration using this approach, it is mainly effective for a single vector (i.e., batch size of 1 in FC layers). However, in practice, generation such as beam search or batched sampling (common in code generation\cite{chen2021evaluatinglargelanguagemodels}) which give higher quality outputs require handling multiple activation vectors simultaneously. This requires overlapping sparse locations across vectors to maintain structured weight sparsity. 

\begin{figure}[H]
    \centering
    \begin{subfigure}[t]{0.57\textwidth}
        \centering
        \includegraphics[width=\textwidth]{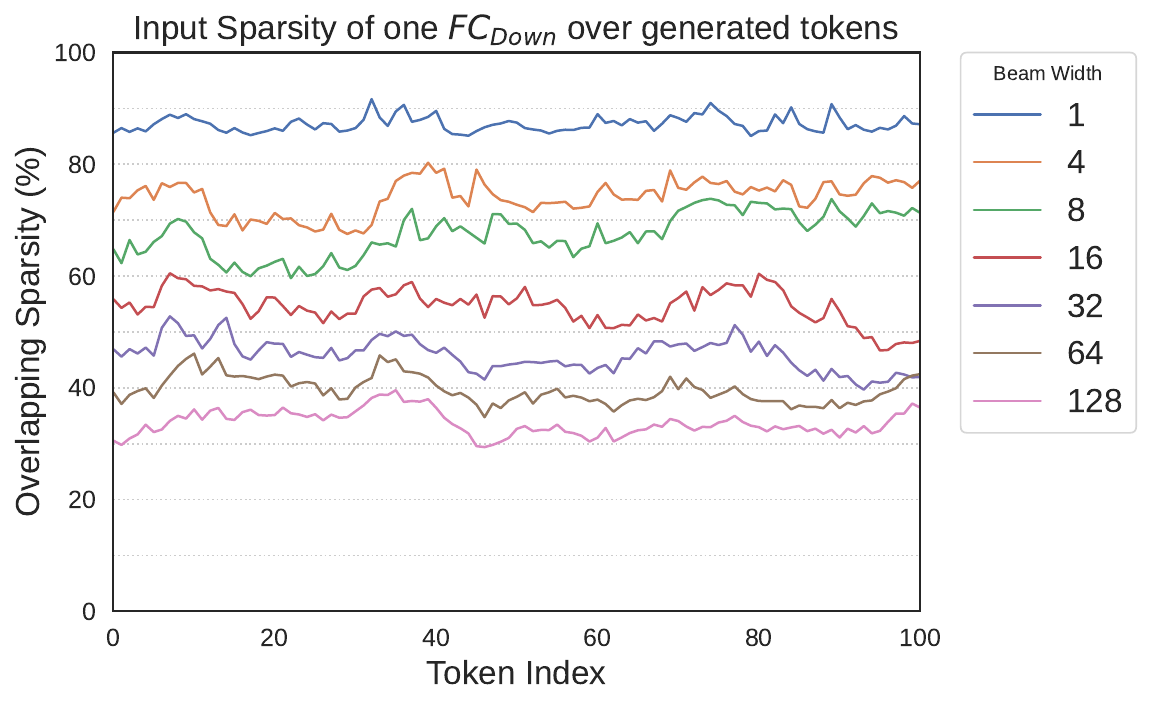}
        \caption{Given a single prompt}
        \label{subfig:overlapping_sparsity_one_prompt}
    \end{subfigure}
    \hfill
    \begin{subfigure}[t]{0.35\textwidth}
        \centering
        \includegraphics[width=\textwidth]{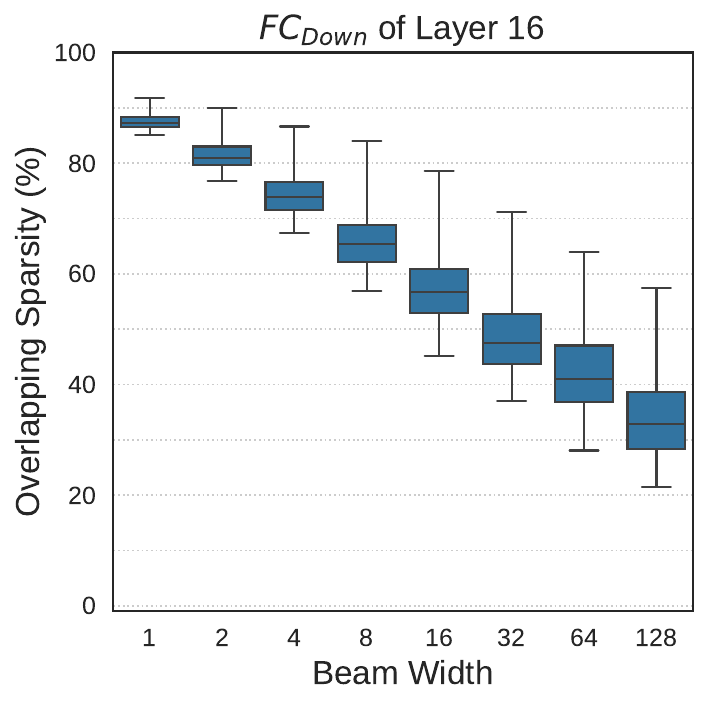} 
        \caption{Averaged over decoding tokens in a dataset}
        \label{subfig:overlapping_sparsity_avg}
    \end{subfigure}
    \caption{Structured Sparsity declines with beam width size (a proxy for high batch input)}
    \label{fig:overlapping_sparsity}
\end{figure}

In our analysis of TurboSparse Mistral 7B, one of the sparsest models achieved through training-based sparsification, sweeping the beam width clearly revealed a decline in overlapping sparsity (see \autoref{fig:overlapping_sparsity}). This issue is further exacerbated in high-throughput serving systems, such as vLLM\cite{kwon2023efficient}, which employs iteration-level batching\cite{yu_orca_2022}. High numbers of parallel batch requests can significantly limit the overall decoding speedup.

On the other hand, the prefill stage of language models and transformer encoders faces similar challenges, if not more pronounced, as their activations consist of multiple vectors. For example, ViT/DeiT3-large \cite{hugo_deit3_2022} with 384x384 image tokenized by a patch size of 16 entails 576 activation vectors at the FC layers. \autoref{tab:many_models} shows that this model can attain up to 59\% of model-wise activation sparsity. While not displayed, we observed that prefill activations exhibit sparsity level similar to those in the decoding phase. Therefore, relying on overlapping sparsity across vectors is leaving a significant amount of sparsity untapped for acceleration. We emphasize the need to address these inference setups to broaden the applicability of acceleration with sparse activation.

\section{Supplementary Experiment Details}
\label{appx:exp_details}
The generic implementation of SCAP is outlined at the beginning of \autoref{sec:experiments}. Our implementation is available at \href{https://github.com/IntelLabs/SCAP}{here}. Essentially, our implementation is primarily based within the Hugging Face ecosystem \cite{Wolf2019TransformersSN}. All pretrained or instruction-tuned models used by SCAP are directly sourced from the \href{https://huggingface.co/models}{model hub}. For task evaluations, we leverage the Language Model Evaluation Harness\cite{eval-harness} for zero-shot tasks\footnote{\label{footnote}Zero-shot tasks refer to the average accuracy across the set of tasks consistent with the evaluation in the CATS paper \cite{lee_cats_2024}, which include WinoGrande, PIQA, SciQ, Hellaswag, BoolQ, Arc-E, and Arc-C.}. When evaluating for the Open LLM leaderboard, we utilize \texttt{LightEval}\cite{lighteval}. 

In terms of compute resources, calibration and sparsification are performed mostly on a single A100-80GB GPU for smaller models and up to 4xA100 GPUs for larger models. For zero-shot task evaluations, we parallelize across more GPUs as needed. Further specific details are provided below.

\subsection{For Section \ref{subsec:pareto}}
\label{appx:pareto}
\href{https://huggingface.co/mistralai/Mistral-7B-v0.1}{\texttt{mistralai/Mistral-7B-v0.1}} and \href{https://huggingface.co/meta-llama/Llama-2-7b-hf}{\texttt{meta-llama/Llama-2-7b-hf}} were the input models used for this study, aligning to CATS. We directly referenced the results reported in \cite{lee_cats_2024}. Pareto fronts of SCAP were constructed based on grid search explained in \autoref{sec:experiments}, with particular task performance, group sparsities presented in \autoref{tab:pareto_mistral} \& \ref{tab:pareto_llama}. Each evaluation was conducted using an identical set of zero-shot tasks\footnotemark[\value{footnote}].

\subsection{For Section \ref{subsec:speedup}, \autoref{tab:generate_speedup}}
\label{appx:details_decoding_speedup}

Our kernel implementation is detailed in \autoref{appx:kernel}. This section provides additional implementation details for \autoref{tab:generate_speedup}, which benchmarked the actual acceleration of the decoding stage achieved through activation sparsity. We profiled the greedy decoding of dense, CATS and SCAP-pruned Mistral-7B in FP32 precision on a single Nvidia L40S GPU with a batch size of 1. This setup was consistent with the original CATS implementation. 

We reproduced the CATS models through our own implementation, ensuring tasks were comparable and observed sparsity was at least equivalent to the target. We then extracted the CATS and SCAP pruning thresholds and proceeded with the benchmarks. Each benchmark consisted of a story segment truncated to the target number of input tokens, requiring the generation of 128 tokens. The reported decoding latency was an average over the generation of the last 127 tokens. For further details, please refer to our \href{https://github.com/IntelLabs/SCAP}{codes}.

\subsection{For Section \ref{subsec:ablate_mode_center}}
\label{appx:details_mode_centering}
We used \href{https://huggingface.co/tiiuae/falcon-7b}{\texttt{tiiuae/falcon-7b}} and \href{https://huggingface.co/mosaicml/mpt-7b}{\texttt{mosaicml/mpt-7b}}, whose FFNs are not GLU-based, with inputs to Down projections originating from GELU that exhibited mode away from zero. For this ablation study, we focused solely on sweeping target sparsity on the inputs of the Down projection, using a grid of 10\%. Evaluations\footnotemark[\value{footnote}] were conducted accordingly.

\subsection{For Section 
\ref{subsec:scap_v_turbosparse}}
\label{appx:details_turbosparse}

TurboSparse \cite{song_turbo_2024}, a SOTA Relufication method, retrofits two ReLUs into pretrained LLMs, as illustrated in \autoref{fig:swiglu_turbosparse}. For our evaluation, we utilized the official \href{https://huggingface.co/PowerInfer/TurboSparse-Mistral-Instruct}{\texttt{TurboSparse-Mistral-Instruct}}, a Relufied version of Mistral-7B that had been uptrained on hundreds of billions of tokens from curated datasets and further fine-tuned for instruction following. 

While it was not possible to align pretraining and instruction datasets perfectly, the closest comparable model was a SCAP on \href{https://huggingface.co/mistralai/Mistral-7B-Instruct-v0.2}{\texttt{Mistral-7B-Instruct-v0.2}}, an instruct fine-tuned version of \texttt{Mistral-7B-v0.2} by MistralAI. We performed a grid search to identify the SCAP-pruned model that maintained overall task performance within a 1\% margin of the baseline average. Both methods were evaluated on the same set of tasks governed by the Open LLM leaderboard using \texttt{LightEval}, with the results and sparsity breakdown provided in \autoref{tab:scap_v_turbosparse}. We observed slight variations in TurboSparse scores compared to those originally reported in the paper.

\subsection{For \autoref{tab:many_models}}
\label{appx:details_conclusion}

\autoref{tab:many_models} lists the models pruned by SCAP to maintain within a -1\% tolerance of their baseline performance, using the grid search outlined in \autoref{sec:experiments}. All input models, except one, were sourced from the Hugging Face \href{https://huggingface.co/models}{model hub}, with hyperlinks provided in the table. The Llama3.1 8B model was locally quantized to 8-bit using data-free symmetrical weight quantization via \href{https://github.com/huggingface/optimum-intel}{Optimum-Intel}. For task evaluation, language models were assessed on zero-shot tasks\footnotemark[\value{footnote}], while Vision Transformers were evaluated using Top-1 accuracy on ImageNet-1k\cite{deng_imagenet_2009}.

The \emph{Sparsity} column records the actual activation sparsity observed during task evaluation, denominated by all targeted FC layers. The last column details the specific FC layers targeted, along with their input sparsity for SCAP calibration. For example, the last row represents a better-trained Vision Transformer\cite{hugo_deit3_2022} with pruning applied to the shared input of QKV, input to Output, Up, and Down projection layers, using SCAP with Mode-Centering. This configuration achieved 59\% activation sparsity during ImageNet-1k evaluation.

\clearpage

\begin{table}[h]
\caption{Activation Sparsity, Zero-Shot Tasks of \textbf{Mistral-7B} pruned by CATS and SCAP}
\label{tab:pareto_mistral}
\centering
\begin{adjustbox}{max width=\textwidth}
{
\renewcommand{\arraystretch}{1.2}
\begin{tabular}{lccccccccccccc}
\toprule
\multirow{2}{*}{\textbf{Model \textbackslash{} Method}} & \multicolumn{4}{c}{\textbf{Input Sparsity of FC Layer}} & \multicolumn{8}{c}{\textbf{Zero-Shot Accuracy (\%)}} \\ 
\cmidrule(lr){2-5} \cmidrule(lr){6-14}
    & Up & Gate & Down & \textbf{FFN} & \textbf{Avg} & \textbf{Relative} & WG & PIQA & SciQ & HS & BoolQ & Arc-E & Arc-C \\ 
\midrule
\textbf{Mistral-7B-v0.1} &  &  &  & 0.0 & 75.3 & Baseline & 74.2 & 80.7 & 95.9 & 61.3 & 83.7 & 80.9 & 50.3 \\ 
\midrule
\rowcolor{cyan}\textbf{CATS 50\%} & 50 & 0 & 50 & \textbf{33.3} & 74.2 & \textbf{-1.5\%} & 72.5 & 80.1 & 94.8 & 61.0 & 81.9 & 78.5 & 50.4 \\ 
\textbf{CATS 70\%} & 70 & 0 & 70 & 46.7 & 72.5 & -3.8\% & 71.9 & 80.0 & 92.9 & 60.6 & 80.3 & 74.9 & 46.9 \\ 
\textbf{CATS 90\%} & 90 & 0 & 90 & 60.0 & 46.9 & -60.6\% & 56.3 & 60.0 & 42.2 & 33.6 & 70.9 & 37.5 & 27.7 \\ 
\midrule
\textbf{SCAP ($s_{up/gate}$: 10\%, $s_{down}$: 50\%)} & 13.2 & 13.2 & 52.7 & 26.4 & 75.3 & 0.0\% & 74.5 & 80.4 & 96.0 & 61.6 & 83.4 & 80.6 & 50.6 \\ 
\textbf{SCAP ($s_{up/gate}$: 10\%, $s_{down}$: 60\%)} & 13.2 & 13.2 & 62.2 & 29.6 & 75.2 & -0.1\% & 74.3 & 80.5 & 95.7 & 61.9 & 82.8 & 80.1 & 51.0 \\ 
\textbf{SCAP ($s_{up/gate}$: 20\%, $s_{down}$: 50\%)} & 22.9 & 22.9 & 52.1 & 32.6 & 75.2 & -0.1\% & 74.0 & 80.5 & 95.7 & 61.5 & 83.9 & 80.5 & 50.3 \\ 
\textbf{SCAP ($s_{up/gate}$: 20\%, $s_{down}$: 60\%)} & 22.9 & 22.9 & 61.7 & 35.8 & 75.0 & -0.3\% & 74.6 & 80.5 & 95.9 & 61.9 & 82.6 & 80.0 & 49.8 \\ 
\textbf{SCAP ($s_{up/gate}$: 35\%, $s_{down}$: 45\%)} & 37.0 & 37.0 & 47.1 & 40.4 & 74.9 & -0.6\% & 74.0 & 80.4 & 95.7 & 60.9 & 83.8 & 80.2 & 49.1 \\ 
\textbf{SCAP ($s_{up/gate}$: 30\%, $s_{down}$: 70\%)} & 32.7 & 32.7 & 71.3 & 45.6 & 74.5 & -1.0\% & 72.7 & 80.3 & 95.8 & 62.0 & 82.4 & 79.5 & 49.2 \\ 
\rowcolor[RGB]{255,127,14} \textbf{SCAP ($s_{up/gate}$: 40\%, $s_{down}$: 60\%)} & 42.0 & 42.0 & 61.7 & \textbf{48.5} & 74.2 & \textbf{-1.5\%} & 71.9 & 80.0 & 95.1 & 60.8 & 82.9 & 79.8 & 48.7 \\ 
\textbf{SCAP ($s_{up/gate}$: 40\%, $s_{down}$: 70\%)} & 42.7 & 42.7 & 71.9 & 52.5 & 72.8 & -3.3\% & 69.1 & 79.8 & 91.8 & 60.1 & 81.7 & 79.2 & 47.7 \\ 
\textbf{SCAP ($s_{up/gate}$: 40\%, $s_{down}$: 80\%)} & 42.8 & 42.8 & 81.4 & 55.7 & 71.6 & -4.8\% & 67.6 & 79.4 & 86.5 & 60.6 & 79.8 & 79.1 & 48.5 \\ 
\textbf{SCAP ($s_{up/gate}$: 50\%, $s_{down}$: 70\%)} & 52.4 & 52.4 & 72.1 & 59.0 & 69.4 & -7.9\% & 70.4 & 79.4 & 75.5 & 56.5 & 77.5 & 78.5 & 47.8 \\ 
\textbf{SCAP ($s_{up/gate}$: 50\%, $s_{down}$: 80\%)} & 52.5 & 52.5 & 81.4 & 62.2 & 67.8 & -9.9\% & 69.3 & 79.4 & 71.5 & 55.3 & 75.4 & 77.7 & 46.5 \\ 
\textbf{SCAP ($s_{up/gate}$: 60\%, $s_{down}$: 70\%)} & 61.9 & 61.9 & 72.0 & 65.3 & 66.7 & -11.5\% & 68.3 & 78.2 & 70.8 & 52.7 & 75.1 & 77.1 & 44.3 \\ 
\bottomrule
\end{tabular}
}
\end{adjustbox}
\end{table}

\vspace{20mm}

\begin{table}[h]
\caption{Activation Sparsity, Zero-Shot Tasks of \textbf{Llama-2-7B} pruned by CATS and SCAP}
\label{tab:pareto_llama}
\centering
\begin{adjustbox}{max width=\textwidth}
{
\renewcommand{\arraystretch}{1.2}
\begin{tabular}{lccccccccccccc}
\toprule
\multirow{2}{*}{\textbf{Model \textbackslash{} Method}} & \multicolumn{4}{c}{\textbf{Input Sparsity of FC Layer}} & \multicolumn{8}{c}{\textbf{Zero-Shot Accuracy (\%)}} \\ 
\cmidrule(lr){2-5} \cmidrule(lr){6-14}
    & Up & Gate & Down & \textbf{FFN} & \textbf{Avg} & \textbf{Relative} & WG & PIQA & SciQ & HS & BoolQ & Arc-E & Arc-C \\ 
\midrule
\textbf{Llama-2-7B} & 0 & 0 & 0 & 0.0 & 70.8 & Baseline & 69.1 & 78.1 & 94.0 & 57.2 & 77.7 & 76.3 & 43.4 \\ 
\midrule
\rowcolor{cyan}\textbf{CATS 50\%} & 50 & 0 & 50 & \textbf{33.3} & 68.9 & \textbf{-2.8\%} & 67.5 & 76.9 & 92.7 & 57.1 & 72.6 & 74.4 & 41.2 \\ 
\textbf{CATS 70\%} & 70 & 0 & 70 & 46.7 & 66.0 & -7.3\% & 66.9 & 75.8 & 90.2 & 55.0 & 65.9 & 70.1 & 38.1 \\ 
\textbf{CATS 90\%} & 90 & 0 & 90 & 60.0 & 51.4 & -37.7\% & 57.4 & 66.3 & 61.1 & 38.5 & 62.8 & 45.7 & 28.1 \\ 
\midrule
\textbf{SCAP ($s_{up/gate}$: 30\%, $s_{down}$: 40\%)} & 32.2 & 32.2 & 42.5 & 35.7 & 70.7 & -0.1\% & 69.7 & 77.6 & 94.0 & 57.0 & 77.5 & 76.2 & 43.1 \\
\textbf{SCAP ($s_{up/gate}$: 20\%, $s_{down}$: 40\%)} & 22.6 & 22.6 & 42.5 & 29.2 & 70.7 & -0.1\% & 68.7 & 77.9 & 93.5 & 57.4 & 77.6 & 76.3 & 43.6 \\
\textbf{SCAP ($s_{up/gate}$: 10\%, $s_{down}$: 50\%)} & 12.9 & 12.9 & 52.3 & 26.1 & 70.7 & -0.1\% & 69.6 & 78.1 & 93.4 & 57.3 & 77.6 & 76.2 & 42.8 \\
\textbf{SCAP ($s_{up/gate}$: 20\%, $s_{down}$: 50\%)} & 22.6 & 22.6 & 52.3 & 32.5 & 70.7 & -0.2\% & 69.2 & 78.1 & 93.7 & 57.3 & 77.3 & 76.5 & 42.6 \\
\textbf{SCAP ($s_{up/gate}$: 30\%, $s_{down}$: 50\%)} & 31.7 & 31.7 & 51.5 & 38.3 & 70.6 & -0.3\% & 70.6 & 77.7 & 93.4 & 57.0 & 77.3 & 75.7 & 42.6 \\
\textbf{SCAP ($s_{up/gate}$: 35\%, $s_{down}$: 50\%)} & 36.5 & 36.5 & 51.5 & 41.5 & 70.3 & -0.7\% & 69.0 & 77.6 & 93.3 & 56.8 & 77.1 & 76.1 & 42.4 \\
\textbf{SCAP ($s_{up/gate}$: 35\%, $s_{down}$: 60\%)} & 36.5 & 36.5 & 61.2 & 44.7 & 70.0 & -1.1\% & 68.4 & 77.3 & 93.5 & 57.0 & 76.4 & 75.1 & 42.5 \\
\textbf{SCAP ($s_{up/gate}$: 40\%, $s_{down}$: 60\%)} & 41.4 & 41.4 & 61.2 & 48.0 & 70.0 & -1.1\% & 68.6 & 77.6 & 93.8 & 56.4 & 76.6 & 74.6 & 42.6 \\
\textbf{SCAP ($s_{up/gate}$: 40\%, $s_{down}$: 70\%)} & 41.9 & 41.9 & 71.5 & 51.8 & 69.6 & -1.7\% & 68.7 & 77.9 & 93.0 & 56.5 & 75.8 & 73.9 & 41.4 \\
\rowcolor[RGB]{255,127,14} \textbf{SCAP ($s_{up/gate}$: 50\%, $s_{down}$: 60\%)} & 51.5 & 51.5 & 61.9 & \textbf{55.0} & 68.8 & \textbf{-2.8\%} & 67.0 & 76.6 & 93.3 & 55.4 & 75.9 & 73.3 & 40.4 \\
\textbf{SCAP ($s_{up/gate}$: 50\%, $s_{down}$: 70\%)} & 51.5 & 51.5 & 71.5 & 58.2 & 68.7 & -3.0\% & 67.3 & 77.0 & 93.6 & 55.1 & 74.8 & 73.3 & 39.8 \\
\textbf{SCAP ($s_{up/gate}$: 50\%, $s_{down}$: 80\%)} & 51.6 & 51.6 & 81.0 & 61.4 & 67.7 & -4.6\% & 64.7 & 75.8 & 92.7 & 55.1 & 73.6 & 71.2 & 40.8 \\
\textbf{SCAP ($s_{up/gate}$: 60\%, $s_{down}$: 70\%)} & 61.2 & 61.2 & 71.4 & 64.6 & 67.2 & -5.4\% & 65.5 & 75.2 & 93.5 & 53.0 & 74.4 & 71.0 & 37.8 \\
\bottomrule
\end{tabular}
}
\end{adjustbox}
\end{table}

\end{document}